\documentclass{article}

 \usepackage[preprint]{neurips_2026}

\usepackage[utf8]{inputenc} % allow utf-8 input
\usepackage[T1]{fontenc}    % use 8-bit T1 fonts
\usepackage{hyperref}       % hyperlinks
\usepackage{url}            % simple URL typesetting
\usepackage{booktabs}       % professional-quality tables
\usepackage{amsfonts}       % blackboard math symbols
\usepackage{nicefrac}       % compact symbols for 1/2, etc.
\usepackage{microtype}      % microtypography
\usepackage{xcolor}         % colors

\usepackage{graphicx}       % \includegraphics
\usepackage{subcaption}     % \begin{subfigure} in appendix figures
\usepackage{amsmath}        % align*, etc.
\usepackage{amssymb}        % \mathfrak in section 3
\usepackage{arydshln}       % \hdashline in result tables
\usepackage{algorithm}      % algorithm float (Appendix B)
\usepackage{algorithmic}    % \STATE/\WHILE pseudocode
\usepackage{amsmath,amsfonts,bm}

\def\eqref#1{equation~\ref{#1}}
\def\1{\bm{1}}

\def\vk{{\bm{k}}}

\def\vo{{\bm{o}}}

\def\vq{{\bm{q}}}

\def\vv{{\bm{v}}}

\def\vx{{\bm{x}}}
\def\vy{{\bm{y}}}

\def\mA{{\bm{A}}}

\def\mK{{\bm{K}}}

\def\mO{{\bm{O}}}

\def\mQ{{\bm{Q}}}

\def\mV{{\bm{V}}}

\DeclareMathAlphabet{\mathsfit}{\encodingdefault}{\sfdefault}{m}{sl}
\SetMathAlphabet{\mathsfit}{bold}{\encodingdefault}{\sfdefault}{bx}{n}

\title{Model-Aware Tokenizer Transfer}

\author{%
  Mykola~Haltiuk\\
  Faculty of Computer Science\\
  AGH University of Krakow\\
  Krakow, Poland\\
  \texttt{mhaltiuk@agh.edu.pl} \\
  \And
  Aleksander~Smywiński-Pohl \\
  Faculty of Computer Science\\
  AGH University of Krakow\\
  Krakow, Poland\\
  \texttt{apohllo@agh.edu.pl} \\
}

\begin{document}

\maketitle

\begin{abstract}
    Large Language Models (LLMs) are trained to support an increasing number of languages, yet their predefined tokenizers remain a bottleneck for adapting models to lower-resource or distinct-script languages. Existing tokenizer transfer methods typically rely on semantic heuristics to initialize new embeddings, ignoring higher-layer model dynamics and limiting transfer quality. We propose Model-Aware Tokenizer Transfer (MATT), a method that incorporates model internals into the tokenizer transfer process. MATT introduces an Attention Influence Modeling (AIM) objective that distills inter-token communication patterns from a source model into a target model with a new tokenizer, providing an efficient warm-up before standard language modeling. Unlike approaches that focus solely on embedding similarity, MATT leverages attention behavior to guide embedding initialization and adaptation. Experiments across diverse linguistic settings show that MATT recovers a large fraction of the original model’s performance within a few GPU hours, outperforming heuristic baselines. These results demonstrate that incorporating model-level signals offers a practical and effective path toward robust tokenizer transfer in multilingual LLMs.\footnote{Implementation is available at \href{https://github.com/Goader/matt}{https://github.com/Goader/matt}.}
\end{abstract}

\section{Introduction}
\label{sec:introduction}
Recent advances in large language models (LLMs) have shifted attention from training monolingual models \citep{jiang2023mistral7b, touvron2023llama} to covering an increasing number of languages \citep{grattafiori2024llama, team2025gemma}. Such multilingual models have become valuable tools for researchers and practitioners working with lower-resource languages. They can be used directly for downstream tasks, help translate English datasets into the target language \citep{rybak2023maupqa}, or act as a robust baseline for further adaptation \citep{ociepa2024bielik}.
Our work focuses on the last scenario: adapting an existing LLM to a new language.

A major practical challenge in this setting is that every pretrained model is tied to a fixed tokenizer. Alternative architectures that avoid a predefined vocabulary, such as the Byte-Latent Transformer \citep{pagnoni-etal-2025-byte} or H-Net \citep{hwang2025dynamic}, are still in the experimental stage and not yet widely adopted. Tokenizers for multilingual models are usually trained to cover many scripts at once and inevitably favor high-resource languages. As a result, languages underrepresented in the tokenizer, most acutely those with distinct alphabets such as Georgian, often receive a very limited share of the vocabulary. This mismatch leads not only to lower accuracy \citep{ali-etal-2024-tokenizer, tamang2024evaluatingtokenizerperformancelarge}, but also to slower processing and inference, which are vital for the end users.

One practical way to mitigate this problem is tokenizer transfer: replacing the original tokenizer of a pretrained model with a new one tailored to the target language and retraining the input and output embeddings \citep{de2020good}. Even models not explicitly trained for multilinguality usually contain some cross-lingual knowledge thanks to shared alphabets or accidental language contamination \citep{blevins-zettlemoyer-2022-language}. Consequently, if we can initialize the new embeddings well, much of the original performance can be recovered and used as a strong starting point for continual pretraining. At the same time, we should not expect this process to introduce entirely new linguistic knowledge, since several studies show that most of the model’s knowledge is stored in the feed-forward layers \citep{dai-etal-2022-knowledge, geva2021transformerfeedforwardlayerskeyvalue, nichani2024understandingfactualrecalltransformers}.

Most existing tokenizer-transfer methods focus almost exclusively on the embedding layer. They construct new embeddings as linear combinations of the original ones, differing mainly in how the combination weights are computed \citep{minixhofer-etal-2022-wechsel, dobler-de-melo-2023-focus, remy2023tik, remy2024trans, li-etal-2025-tokalign}. By ignoring the higher layers, these approaches overlook how the model actually processes tokens. More recent work, such as Zero-Shot Tokenizer Transfer by \citet{NEURIPS2024_532ce4fc}, leverages the full model by training a hypernetwork with a language modeling objective to predict embeddings. While effective, this strategy is computationally demanding because language modeling requires full forward and backward passes through the model.

To overcome these limitations, we introduce \textbf{Model-Aware Tokenizer Transfer (MATT)}, a method that leverages the internal behavior of the pretrained model rather than relying only on surface semantics. At the core of MATT is \textbf{Attention Influence Modeling (AIM)} objective.

AIM encourages the model with the new tokenizer to reproduce the inter-token interactions of the original model’s attention layers. In effect, the original model acts as a teacher, while the model with the new tokenizer serves as a student that learns to match its attention patterns. This procedure distills structural knowledge about token relationships directly from the teacher, providing a richer and more informative initialization than relying on an embedding layer alone.

MATT is orthogonal to existing heuristics based on semantic similarity and can be combined with them. It acts as an efficient warm-up stage before conventional language model pretraining: a small residual gap to the original model, most visible on generative tasks, typically remains, but MATT closes most of it at a fraction of the cost of language-modeling-based alternatives, leaving subsequent continual pretraining as the standard path to fully close the gap.

We evaluate MATT by transferring the tokenizers of Gemma 3 \citep{team2025gemma} and Qwen 3 \citep{qwen3technicalreport} models to extended versions that increase compression and expand coverage for several languages, including English, German, Japanese, Arabic, Swahili, and Ukrainian. Across multiple settings, MATT consistently recovers a substantial portion of the original model’s performance on both generative and discriminative tasks, while requiring only a few GPU hours and outperforming heuristic-based transfer methods.

Our contributions can be summarized as follows:

\begin{itemize}
    \item Attention Influence Modeling (AIM): a novel distillation objective that aligns the attention dynamics of two models with different tokenizers.
    \item Model-Aware Tokenizer Transfer (MATT): an efficient tokenizer-transfer method that exploits model dynamics instead of relying solely on semantic relationships, outperforming heuristic and optimization-based baselines at substantially lower computational cost than language modeling objectives.
    \item Comprehensive evaluation: experiments across multiple languages and models demonstrating the effectiveness and efficiency of MATT.
\end{itemize}

\section{Related work}
\label{sec:related}
\paragraph{Large language models and vocabulary size}

Large Language Models are becoming increasingly multilingual. Early open-source models focused almost exclusively on English \citep{jiang2023mistral7b, touvron2023llama, almazrouei2023falcon}, but most recent releases include at least several languages and offer partial support for many more. This shift toward multilinguality has changed how researchers choose vocabulary size.

Studies show that larger vocabularies can improve model quality \citep{takase-etal-2025-large, liang-etal-2023-xlm}, but they also slow training and inference. As a result, most current foundation models use vocabularies of about 100 to 250 thousand tokens, with strongly multilingual models leaning toward the upper end. This sweet spot, first popularized by XLM-RoBERTa \citep{conneau-etal-2020-unsupervised}, continues in more recent models such as Gemma \citep{team2025gemma}, Aya Expanse \citep{dang2024aya}, and even GPT-5.\footnote{\href{https://github.com/openai/tiktoken}{https://github.com/openai/tiktoken}} Going beyond this range rarely pays off: performance gains are small, and efficiency drops sharply. As a result, tokenizers cannot achieve an optimal compression rate for every language, creating a need for techniques that allow efficient transfer of tokenizers to specific languages or domains without requiring very large vocabularies.

\paragraph{Heuristics-based embedding initialization methods}

When transferring a tokenizer to a new language or domain, the main challenge is initializing embeddings for tokens that did not exist in the original model. Early work on tokenizer transfer \citep{artetxe-etal-2020-cross, gogoulou-etal-2022-cross, de2020good} focused on proving that transfer was possible, so embedding initialization received little attention. Simple strategies were used, including random initialization, taking the mean of existing embeddings, sampling from their distribution, copying the embedding of a random token, or using token frequency as a guide.

Later research began to exploit semantic relationships between tokens. WECHSEL \citep{minixhofer-etal-2022-wechsel} was an influential step: it trained FastText \citep{bojanowski2017enriching} embeddings for the source and target languages and used a translation vocabulary to identify the closest source tokens for each new token. New embeddings were then initialized as weighted averages of these source embeddings.
Several methods followed a similar direction. OFA \citep{liu-etal-2024-ofa} and Tik-to-Tok \citep{remy2023tik} refined the idea of using cross-lingual similarities, while Transtokenization \citep{remy2024trans} created its own token-level translation dictionary with FastAlign \citep{dyer2013simple}. Hyper-OFA \citep{ozeren-etal-2025-hyperofa} went further by training a hypernetwork to map tokens from an external multilingual space into the model’s embedding space, avoiding the need for simplistic linear combinations. TokAlign \citep{li-etal-2025-tokalign} took a co-occurrence perspective, training two GloVe \citep{pennington-etal-2014-glove} models on the same corpus to learn a one-to-one alignment matrix between tokens.

As LLMs became more multilingual, overlap between source and target vocabularies became an important resource. FOCUS \citep{dobler-de-melo-2023-focus} trains a FastText model on text tokenized with the target vocabulary, then initializes new embeddings as similarity-weighted averages of overlapping tokens. CLP Transfer \citep{ostendorff2023efficient} takes advantage of topological similarities of the latent space across model sizes within the same family: embeddings are first trained on a smaller related model and then aligned to the target model by measuring similarities with overlapping tokens.

\paragraph{Beyond heuristics}

While heuristics provide a practical starting point, they have limitations. An alternative is to train new embeddings directly by continuing language modeling with all other parameters frozen \citep{de2020good}, but this is computationally costly.

Another approach by \citet{NEURIPS2024_532ce4fc} trains a universal hypernetwork for a given language model by sampling tokenizers from a diverse distribution during the language modeling stage. Once the hypernetwork is trained, we can initialize embeddings for various tokenizers effortlessly, achieving a solid baseline for further continual pretraining. However, training such a hypernetwork is a compute-heavy task, requiring forward and backward passes through the whole model in every step to update the hypernetwork weights, limiting its practicality in settings where we already have a defined target tokenizer and the trained hypernetwork is not available beforehand.

More recently, Token Distillation \citep{dobler2026token} optimizes the embedding of each new token to produce, when fed through the model, the same hidden state that the original model yields for the equivalent multi-token sequence.

\section{Method}
\label{sec:method}
\subsection{Intuition}
\label{subsec:intuition}

Large Language Models generate text one token at a time. Decoder-only transformers, which form the backbone of most modern LLMs, follow the following steps: the embedding of the most recently generated token is passed through a stack of attention and feed-forward layers, and finally projected by the LM head to produce a probability distribution over the next token.

Assuming the input embedding of the last token is correct, the feed-forward layers will not damage its representation. The main source of potential distortion lies in the attention layers, where each token interacts with the context. Changing the tokenizer introduces new tokens into the context, altering these interactions and thus the internal representations that drive next-token prediction. Our goal is to train a model using a new tokenizer so that, despite these changes, its attention layers produce output embeddings similar to those generated by the original tokenizer.

\subsection{Prerequisites}

Consider an input string $s$ and a tokenization function $T$, which produces a token sequence $T(s) = (t_1, t_2, \dots, t_n)$ of length $n$.

In each attention layer,\footnote{While we proceed with a single-head definition, it is directly applicable to multi-head, multi-query, or grouped-query attention variants.} the inputs are the query ($\bm{Q}$), key ($\mK$), and value ($\mV$) state matrices, producing the output state matrix ($\mO$). Each of these can be seen as a collection of vector states for every token $t_i$:
\begin{equation}
    \label{eq:state-matrices}
    \mQ = \left[\begin{array}{c}\vq_1 \\ \vq_2 \\ \dots \\ \vq_n\end{array}\right]_{n \times h},\quad
    \mK = \left[\begin{array}{c}\vk_1 \\ \vk_2 \\ \dots \\ \vk_n\end{array}\right]_{n \times h},\quad
    \mV = \left[\begin{array}{c}\vv_1 \\ \vv_2 \\ \dots \\ \vv_n\end{array}\right]_{n \times h},\quad
    \mO = \left[\begin{array}{c}\vo_1 \\ \vo_2 \\ \dots \\ \vo_n\end{array}\right]_{n \times h},
\end{equation}

where $h$ is the hidden size.

Attention is computed as:
\begin{equation}
    \label{eq:attention}
    \mO = \text{Attention}(\mQ, \mK, \mV) = \text{softmax}\left(\frac{\mQ\mK^T}{\sqrt{d_k}}\right)\mV = \mA\mV,
\end{equation}
where $\mA$ is the attention matrix of shape $n \times n$, that contains weights with which the value states are aggregated into the output state.

We can break down the final matrix multiplication $\mA\mV$ into a chain of value states ($\mV$) averages for each token, weighted by the attention matrix $\mA$. The output state for the token $t_i$ would then look the following way:
\begin{equation}
    \label{eq:per-token-output}
    \vo_i = \text{softmax}\left(\frac{\vq_i \mK^T}{\sqrt{d_k}}\right)\mV = \mA_{i,:}\mV = \sum_{j=1}^n \mA_{i,j} \vv_j = \sum_{j=1}^n \vv^*_{i,j},
\end{equation}
where $\vv^*_{i,j} = \mA_{i,j}\vv_j$ is a weighted value state for the token $t_j$ given the query token $t_i$.

\subsection{Segment-level interpretation of attention}

To compare attention outputs across different tokenizers, we introduce a segmentation function $S$ that splits the input string $s$ into segments $(s_1, s_2, \dots, s_m)$ while respecting a set of tokenization functions $\mathcal{T}$:
\begin{equation}
    \label{eq:segmentation}
    \begin{gathered}
        S(s;\,\mathcal{T}) = (s_1, s_2, \dots, s_m), \, \text{such that} \\
        \forall\,T \in \mathcal{T}:\,T(s_1) \circ T(s_2) \circ \cdots \circ T(s_m) = T(s),
    \end{gathered}
\end{equation}
where $\circ$ is a concatenation operator. This ensures that no segment boundary lies within any token produced by any tokenization function in $\mathcal{T}$.

The most intuitive approach is a function that splits the input string into words, and the rest of the section is explained in relation to this function. However, for some languages, word segmentation can be ambiguous; thus, in practice, we define our segmentation function to always choose segments of minimal length that still satisfy the above condition (see Appendix~\ref{apx:segmentation} for the algorithm).

Given $S$, we define weighted value states for a segment $s_k$ with respect to a query token $t_i$:
\begin{equation}
    \label{eq:segment-value}
    \bm{\mathfrak{s}}_{i, k} = \sum_{\{j\,:\, t_j \in T(s_k)\}} \vv^*_{i, j}.
\end{equation}

The output state for token $t_i$ can then be expressed as a sum over segments:
\begin{equation}
    \label{eq:output-segment-sum}
    \vo_i = \sum_{j=1}^n \vv^*_{i,j} = \sum_{j=1}^m \bm{\mathfrak{s}}_{i,j}.
\end{equation}

To move from token-level to segment-level interpretation, we replace individual query tokens with segment representations.
Since the output state of each token is designed to predict the next token, it is natural to require that the output state of a segment should similarly carry enough information to predict the next segment.
Because the language modeling head still operates at the token level, we approximate "predicting the next segment" by predicting the first token of that next segment.

Consider a segment $s_i$ whose tokens are $T(s_i) = (t_a, t_{a+1}, \dots, t_b)$. The final token $t_b$ produces the output state used to generate the next token $t_{b+1}$, which begins the following segment $s_{i+1}$. We therefore define a function $\ell_T$ that maps a segment index to the index of its last token:

\begin{equation}
    \label{eq:last-token-index}
    \ell_T(i) = b.
\end{equation}

The query state of segment $s_i$ is set equal to the query state of its last token:

\begin{equation}
    \label{eq:segment-query}
    \bm{\mathfrak{q}}_i = \vq_{\ell_T(i)} = \vq_b,
\end{equation}

and the output state of the segment is computed from this query state:
\begin{equation}
    \label{eq:segment-output}
    \bm{\mathfrak{o}}_i = \text{softmax}\left(\frac{\bm{\mathfrak{q}}_i \mK^T}{\sqrt{d_k}}\right)\mV = \text{softmax}\left(\frac{\vq_{\ell_T(i)} \mK^T}{\sqrt{d_k}}\right)\mV = \vo_{\ell_T(i)} = \sum_{j=1}^m \bm{\mathfrak{s}}_{\ell_T(i),j}.
\end{equation}

\subsection{Attention Influence Modeling}

As described in the Section~\ref{subsec:intuition}, our goal is to train the model with a new tokenizer $T^{\prime}$ so that its output states match those of the original model with tokenizer $T$.

Since we can enforce a common segmentation function $S$, we approximate this by requiring the new model to produce the same segment-level outputs $\bm{\mathfrak{o}}^{\prime}_i$ as the old ones -- $\bm{\mathfrak{o}}_i$. A more detailed objective also matches the weighted value states $\bm{\mathfrak{s}}_{\ell_T(i),j}$ and $\bm{\mathfrak{s}}^{\prime}_{\ell_{T^{\prime}}(i),j}$ of every segment $s_j$ for each query segment $s_i$, with the causal constraint $j \le i$.

Given the above, we define the \textbf{Attention Influence Modeling} objectives (normal and simplified):
\begin{equation}
    \label{eq:aim}
    \mathcal{L}_{AIM} = \frac{2}{m(m+1)} \sum_{i=1}^m \sum_{j=1}^i \mathcal{L}^*(\bm{\mathfrak{s}}_{\ell_T(i), j}, \bm{\mathfrak{s}}^{\prime}_{\ell_{T^{\prime}}(i), j}),
\end{equation}
\begin{equation}
    \label{eq:aim-star}
    \mathcal{L}_{AIM^*} = \frac{1}{m} \sum_{i=1}^m \mathcal{L}^*(\bm{\mathfrak{o}}_{i}, \bm{\mathfrak{o}}^{\prime}_{i}),
\end{equation}
where $\mathcal{L}^*(\vx, \vy)$ can be any loss function, that brings $\vx$ and $\vy$ closer. In Section~\ref{sec:experiments}, we experiment with MSE and Cosine Embedding losses.

\begin{figure*}[h]
    \centering
    \includegraphics[width=\linewidth]{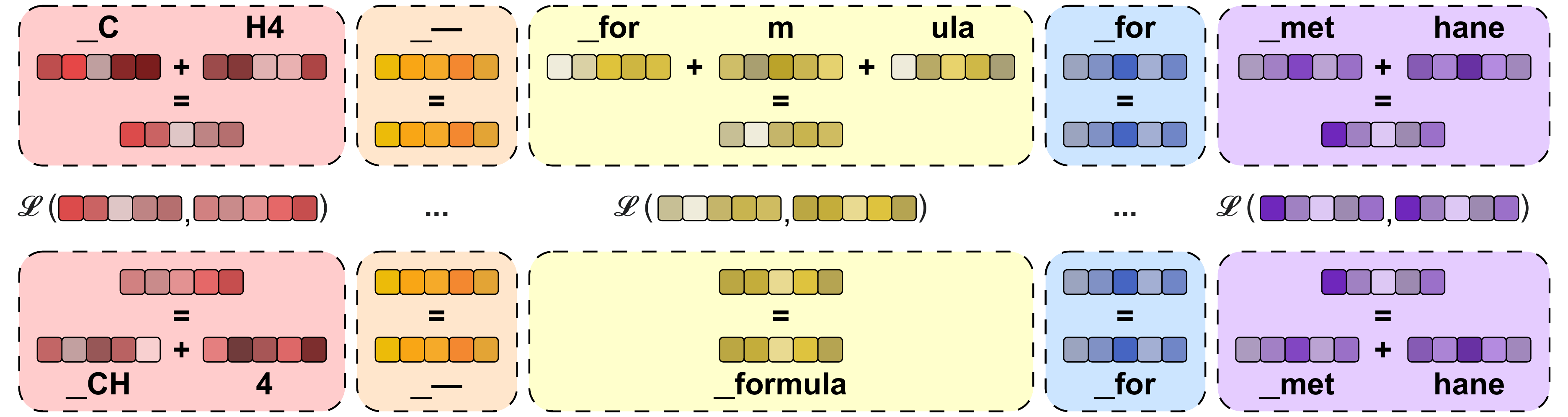}
    \caption{Attention Influence Modeling (AIM) objective with word segmentation. For each input, the weighted value vectors $\vv^*_{i,j}$ of the original tokens $t_j$ are aggregated into segment-level vectors $\bm{\mathfrak{s}}_{i, k}$ according to a chosen word-segmentation function. The model trained with the new tokenizer produces its own segment representations $\bm{\mathfrak{s}}^{\prime}_{i, k}$. The AIM objective encourages these new segment representations to stay close to the segment representations $\bm{\mathfrak{s}}_{i, k}$ computed from the model using the old tokenizer. All this happens with respect to the query state $\bm{\mathfrak{q}}_5$ of the 5th segment (\texttt{\_methane}), which is equal to the query state of its last token -- \texttt{hane}.}
    \label{fig:aim}
\end{figure*}

Figure~\ref{fig:aim} illustrates an example of applying AIM to the text \textit{CH4 – formula for methane}. In this case, we use a word segmentation function together with different tokenization functions for a given query state $\bm{\mathfrak{q}}_5$, where the segment $s_5$ corresponds to \texttt{\_methane}.

Appendix~\ref{apx:aim-visualization} complements Figure~\ref{fig:aim} with a sequence-level view of the same alignment, plotting the weighted value states $\vv^*_{i, j}$ across the full causal attention matrix with segments highlighted; these segment-level representations are matched and optimized to be equal under the $\mathcal{L}_{AIM}$ loss.

\subsection{Technical details}
\label{subsec:technical-details}

During training, the model with the old tokenizer $T$ is kept frozen. The model with the new tokenizer $T^{\prime}$ has all layers frozen except the input embeddings. As a small modification to the basic training setup, we partially freeze the embedding matrix: tokens that are shared between the old and new tokenizers are initialized from the original model and kept fixed, while only the embeddings of new, non-overlapping tokens are updated during training.

To speed up convergence, we initialize new embeddings using FOCUS \citep{dobler-de-melo-2023-focus}. We train with AdamW \citep{DBLP:journals/corr/abs-1711-05101}, a constant learning rate of $1 \times 10^{-4}$, and no weight decay. However, it should be noted that we have not performed extensive hyperparameter tuning, so using learning rate scheduling, adapting the learning rate, weight decay, and other hyperparameters may yield significantly better results.

MATT offers a key advantage over standard language modeling with frozen non-embedding parameters: greater efficiency.
Since AIM is defined at the attention-layer level, we can decide how much of the model to include in the tokenizer transfer by selecting the layer depth at which AIM is applied. Specifically, by choosing a value of $n$, we take only the first $n$ layers into account. This allows us to balance efficiency and performance. 

We ablate the choice of MATT target layer in Appendix~\ref{apx:ablation-studies} and find a consistent pattern: performance climbs through the early layers, plateaus across the middle of the model, and slightly degrades on the final layers -- mirroring the trend reported for Token Distillation~\cite{dobler2026token}. The plateau aligns with the formation of coherent word-level representations as tokens are detokenized in early and middle layers~\citep{kaplan2024tokens}. Because training time and VRAM grow linearly with target depth, we recommend a layer near the start of this plateau, empirically around the first quarter to first third of the model, and use this rule throughout (e.g.\ layer 12 of 34 for Gemma 3 12B PT in Section~\ref{subsec:main-results}).

% Since only input embeddings are trained, tied input–output embeddings are advantageous, as the tuned input embeddings can be reused in the LM head. Models without tied embeddings still benefit from input tuning, but to a significantly lesser extent; handling untied settings is left for future work.

% updated

Since only input embeddings are trained with AIM, tied input–output embeddings are advantageous, as the tuned input embeddings can be reused in the LM head. For models without tied embeddings, we combine AIM with an NTP loss to ensure proper optimization of the output embeddings, at a slight cost in training efficiency; this combined variant is used \emph{only} in the untied-embeddings setting (Appendix~\ref{apx:untied-embeddings}). Throughout the rest of the paper, ``MATT'' refers to the AIM objective alone.

\section{Experiments}
\label{sec:experiments}
We conducted a series of experiments across different languages, model families, and scales to evaluate 
the effectiveness of the MATT method compared to existing heuristic- and optimization-based approaches. 
In each experiment, we first trained a tokenizer with a higher compression rate than the original one, merged it with the base tokenizer, and then applied tokenizer transfer to the extended vocabulary.
We have chosen Ukrainian as a language with Cyrillic alphabet that is
not well represented in the dictionaries of the major  LLMs and on the other hand that is included in 
multiple benchmarks, allowing for the inspection of the method’s performance in different scenarios.
For the multilingual setting we adapt to four typologically diverse target languages -- Arabic, German, Japanese, and Swahili, which vary in resource availability, writing system, and language family.

Additional experiments, including convergence speed tests (Appendix~\ref{apx:convergence-speed}) and ablation studies (Appendix~\ref{apx:ablation-studies}), are presented to complement the main results.

\subsection{Main results}
\label{subsec:main-results}

Our primary evaluation uses the Gemma 3 12B PT model \citep{team2025gemma}. We replaced its default tokenizer with an extended version that improves Ukrainian coverage, raising the compression rate from 2.98 to 4.44. This increase translates to an almost 50\% speedup during inference. We compare the following methods:

\begin{itemize}
    \item \textbf{WECHSEL} -- transfer using the English–Ukrainian vocabulary from the official implementation.\footnote{\href{https://github.com/CPJKU/wechsel}{https://github.com/CPJKU/wechsel}}

    \item \textbf{Transtokenizers} -- token alignment via FastAlign using parallel corpora (OpenSubtitles \citep{lison-tiedemann-2016-opensubtitles2016} and NLLB \citep{nllb2022}) and the official \texttt{transtokenizers}\footnote{\href{https://github.com/LAGoM-NLP/transtokenizer}{https://github.com/LAGoM-NLP/transtokenizer}} toolkit.

    \item \textbf{TokAlign} -- GloVe embeddings trained on 2 million Ukrainian documents (approximately 1.86 billion Gemma tokens) from the Kobza corpus \citep{haltiuk-smywinski-pohl-2025-path}, used to create a one-to-one alignment matrix with the official implementation.\footnote{\href{https://github.com/ZNLP/TokAlign}{https://github.com/ZNLP/TokAlign}}

    \item \textbf{FOCUS} -- FastText embeddings trained on the same data as TokAlign, with initialization performed via the \texttt{deepfocus}\footnote{\href{https://github.com/konstantinjdobler/focus}{https://github.com/konstantinjdobler/focus}} package.

    \item \textbf{NTP} -- initialized with one of the above methods, and trained using the Next Token Prediction (NTP) objective with non-embedding layers frozen. We compare several versions of this baseline corresponding to 50\%, 100\%, and 150\% of the training budget dedicated to MATT.

    \item \textbf{Token Distillation} -- trained using the \texttt{token-distillation}\footnote{\href{https://github.com/konstantinjdobler/token-distillation}{https://github.com/konstantinjdobler/token-distillation}} package. The number of mined examples per token was set to match MATT's compute budget.

    \item \textbf{MATT} -- initialized with FOCUS embeddings and trained on around 240 million Ukrainian tokens from Kobza using the AIM objective with MSE loss on the 12th layer out of 34, original embeddings are frozen, and all other hyperparameters remain unchanged (see Section~\ref{subsec:technical-details}).

\end{itemize}

All evaluations in this section are restricted to Ukrainian. Specifically, we use the Ukrainian subset of Belebele \citep{bandarkar-etal-2024-belebele}, Global MMLU \citep{singh-etal-2025-global}, and XL-SUM \citep{hasan-etal-2021-xl}; for translation we use both the en$\rightarrow$uk and uk$\rightarrow$en directions of Long FLORES \citep{paniv:2025:RANLP}, a document-level extension of FLORES \citep{nllb2022, flores-101, low-resource-machine-translation}, and the en$\rightarrow$uk direction of WMT24++ \citep{deutsch2025wmt24expandinglanguagecoverage}. Including both Long FLORES directions probes generative performance into and out of the target language. Evaluation is performed with the \texttt{lm-evaluation-harness} framework \citep{eval-harness} with a 3-shot prompt.

\renewcommand{\arraystretch}{1.2}
\begin{table*}[h]
    \caption{Performance of Gemma 3 12B PT model with different tokenizer transfer methods on Belebele and Global MMLU (accuracy, \%), Long FLORES, WMT, and XL-SUM (BLEU). The "Avg Disc" column reports the average of Belebele and Global MMLU scores, as well as "Avg Gen" -- of Long FLORES (both directions), WMT, and XL-Sum.}
    \resizebox{\linewidth}{!}{%
        \centering
        \label{table:main-results}
        \begin{tabular}{l|c|cccccc|cc}

            \hline
                \textbf{Model}
                & \begin{tabular}{@{}c@{}}\textbf{Training} \\ \textbf{Time}\end{tabular}
                & \textbf{Belebele}
                & \begin{tabular}{@{}c@{}}\textbf{Global} \\ \textbf{MMLU}\end{tabular}
                & \multicolumn{2}{c}{\begin{tabular}{@{}cc@{}}\multicolumn{2}{c}{\textbf{Long FLORES}} \\ \textbf{en$\rightarrow$uk} & \textbf{uk$\rightarrow$en}\end{tabular}}
                & \textbf{WMT}
                & \textbf{XL-Sum}
                & \begin{tabular}{@{}c@{}}\textbf{Avg} \\ \textbf{Disc}\end{tabular}
                & \begin{tabular}{@{}c@{}}\textbf{Avg} \\ \textbf{Gen}\end{tabular}
            \\
            % gemma
            \hline
            Gemma 3 12B PT & - & 89.33 & 67.03 & 14.36 & 26.32 & 3.52 & 6.52 & 78.18 & 12.68
            \\
            \hdashline
            \multicolumn{10}{c}{\textbf{Heuristics}}
            \\
            \hdashline
            WECHSEL & - & 22.67 & 24.61 & 0.00 & 0.00 & 0.00 & 0.00 & 23.64 & 0.00
            \\
            Transtokenizers & - & 61.89 & 46.03 & 0.04 & 0.10 & 0.09 & 0.02 & 53.96 & 0.06
            \\
            TokAlign & - & 31.44 & 32.98 & 0.00 & 0.31 & 0.00 & 0.01 & 32.21 & 0.08
            \\
            FOCUS & - & 48.78 & 37.14 & 1.01 & 6.19 & 0.88 & 0.20 & 42.96 & 2.07
            \\
            \hdashline
            \multicolumn{10}{c}{\textbf{Optimization Based}}
            \\
            \hdashline
            Transtokenizers w/ NTP & 3h 30m & 82.44 & 59.02 & 3.64 & 1.24 & 0.88 & 4.06 & 70.73 & 2.46
            \\
            Transtokenizers w/ NTP & 7h 00m & 85.22 & 59.83 & 4.63 & 2.17 & 0.95 & 4.80 & 72.53 & 3.14
            \\
            Transtokenizers w/ NTP & 10h 30m & 85.67 & 59.38 & 5.13 & 2.07 & 0.96 & 4.80 & 72.53 & 3.24
            \\
            FOCUS w/ NTP & 3h 30m & 85.44 & 57.38 & 3.51 & 20.26 & 2.13 & 4.32 & 71.41 & 7.56
            \\
            FOCUS w/ NTP & 7h 00m & 87.00 & 60.55 & 4.32 & 23.00 & 2.51 & 5.04 & 73.78 & 8.72
            \\
            FOCUS w/ NTP & 10h 30m & 87.44 & 60.57 & 4.34 & 21.73 & 2.60 & 5.16 & 74.01 & 8.46
            \\
            Token Distillation & 5h 30m & 82.00 & 58.10 & 0.00 & 21.54 & 0.41 & 3.71 & 70.05 & 6.42
            \\
            Token Distillation & 8h 20m & 82.89 & 59.23 & 0.00 & 21.92 & 0.72 & 4.06 & 71.06 & 6.68
            \\
            MATT & 7h 00m &\textbf{89.56} & \textbf{64.98} & \textbf{8.70} & \textbf{27.89} & \textbf{4.71} & \textbf{5.95} & \textbf{77.27} & \textbf{11.81}
            \\
            \hline
        \end{tabular}

    }
\end{table*}

Table~\ref{table:main-results} shows a clear advantage of MATT over all other methods. While heuristic-based approaches such as FOCUS and Transtokenizers can regain up to about 70\% of the original model’s accuracy on discriminative tasks, they reach no more than about 16\% of the original generative performance.

Optimization-based approaches show substantially stronger recovery. NTP yields consistent results across different heuristic initializations: in its best configuration, it recovers nearly 95\% of the original model’s discriminative performance and around 70\% of its generative capabilities. Token Distillation matches NTP on the discriminative side (around 91\%) but its generative behavior is sharply asymmetric: it preserves uk$\rightarrow$en translation at near-original quality (21.92 BLEU vs 26.32), yet collapses to near-zero on en$\rightarrow$uk and on WMT (en$\rightarrow$uk). MATT, by contrast, achieves over 90\% of the original generative performance with balanced quality across both translation directions, while keeping discriminative accuracy close to the unmodified model. These results demonstrate the advantage of the model-aware approach to tokenizer transfer, particularly given the extremely low computational costs required.

% The most interesting is the minimal improvements achieved by NTP when comparing 100\% and 150\% compute budgets (7h and 10.5h of training time, respectively). The NTP training quickly saturates, and MATT’s performance does not seem to be achievable within a reasonable budget. We do not continue with further training due to the limited computational budget.

% As for the MATT, it also saturates quickly (see Appendix~\ref{apx:convergence-speed}), but at a higher performance level. We speculate that further improvements are difficult to achieve with mere embedding training, and thus require unfreezing the model’s layers. This is largely due to the number of newly introduced tokens (over 80,000), which inevitably changes the model’s dynamics, and would benefit from full fine-tuning. Unfortunately, we have not yet experimented with continual pretraining after MATT due to limited computational budget.

Both optimization-based methods saturate quickly. NTP shows minimal improvement between 100\% and 150\% of MATT’s compute budget (7h vs 10.5h), and MATT-level performance does not appear attainable within a reasonable budget. MATT also saturates (Appendix~\ref{apx:convergence-speed}), but at a substantially higher level; further gains likely require unfreezing the model’s layers, since the over-80{,}000 newly introduced tokens meaningfully alter model dynamics. We have run only limited continual-pretraining experiments after MATT due to computational constraints, and a systematic comparison of embedding-only adaptation and full-model fine-tuning is left for future work.

% Interestingly, the results for the instruction-tuned model suggest that MSE loss may be more beneficial for generative tasks than cosine embedding loss, recovering over 70\% of the original model's performance.

\subsection{Multilingual results}
\label{subsec:multilingual-results}

In the multilingual setting, we apply MATT independently per target language to two open mid-scale models, Gemma~3 4B PT \citep{team2025gemma} and Qwen~3 4B \citep{qwen3technicalreport}, on four typologically diverse target languages: Arabic, German, Japanese, and Swahili. For each (model, target language) pair we construct a per-language extended tokenizer that improves compression for the target language while preserving the original tokenizer's coverage of other languages, and we evaluate MATT, FOCUS, and Token Distillation against the unmodified Baseline. Per-language vocabulary sizes and compression rates are reported in Appendix~\ref{apx:multilingual-detail} (Table~\ref{table:multilingual-tokenizer-stats}); the extended tokenizers improve compression on the target language by 1.2$\times$ to 1.6$\times$ over the base. Training data is drawn from HPLT 2.0 Cleaned \citep{burchell2025expanded}, with approximately 250 million tokens per language. Following the rule of thumb in Section~\ref{subsec:technical-details}, AIM is applied at the 12th layer for Gemma~3 4B PT (out of 34) and the 16th layer for Qwen~3 4B (out of 36); all other hyperparameters match Section~\ref{subsec:main-results}. Token Distillation training was configured to match or marginally exceed MATT's per-language compute budget on the same H100 (see the Time column of Table~\ref{table:multilingual-detail}), so MATT's lead over Token Distillation cannot be attributed to having received more training time. Performance is reported on the target-language subsets of Belebele, Global MMLU, MMMLU \citep{hendrycks2020measuring}, and XL-Sum, and on the en$\rightarrow$x and x$\rightarrow$en directions of Long FLORES.

\renewcommand{\arraystretch}{1.2}
\begin{table*}[h]
    \caption{Multilingual results, averaged across four target languages (Arabic, German, Japanese, Swahili). Each column is the mean of the per-language values from Appendix~\ref{apx:multilingual-detail}; Belebele, Global MMLU, and MMMLU report accuracy (\%); Long FLORES (both directions) and XL-Sum report BLEU. ``Avg Disc'' averages the three discriminative metrics; ``Avg Gen'' the three generative ones. XL-Sum has no German subset, so its mean is over Arabic, Japanese, and Swahili.}
    \resizebox{\linewidth}{!}{%
        \centering
        \label{table:multilingual-results}
        \begin{tabular}{l|cccccc|cc}
            \hline
                \textbf{Method}
                & \textbf{Belebele}
                & \begin{tabular}{@{}c@{}}\textbf{Global} \\ \textbf{MMLU}\end{tabular}
                & \textbf{MMMLU}
                & \multicolumn{2}{c}{\begin{tabular}{@{}cc@{}}\multicolumn{2}{c}{\textbf{Long FLORES}} \\ \textbf{en$\rightarrow$x} & \textbf{x$\rightarrow$en}\end{tabular}}
                & \textbf{XL-Sum}
                & \begin{tabular}{@{}c@{}}\textbf{Avg} \\ \textbf{Disc}\end{tabular}
                & \begin{tabular}{@{}c@{}}\textbf{Avg} \\ \textbf{Gen}\end{tabular}
            \\
            \hline
            \multicolumn{9}{c}{\textbf{Gemma 3 4B PT}} \\
            \hdashline
            Baseline & 70.53 & 55.00 & 49.26 & 7.93 & 20.22 & 4.14 & 58.26 & 10.76 \\
            FOCUS & 43.33 & 40.50 & 36.63 & 1.93 & 8.39 & 0.18 & 40.15 & 3.50 \\
            Token Distillation & 57.39 & 45.81 & 43.52 & 1.31 & 12.87 & 1.09 & 48.91 & 5.09 \\
            MATT & \textbf{66.03} & \textbf{52.81} & \textbf{46.84} & \textbf{3.94} & \textbf{13.22} & \textbf{1.27} & \textbf{55.23} & \textbf{6.14} \\
            \hdashline
            \multicolumn{9}{c}{\textbf{Qwen 3 4B}} \\
            \hdashline
            Baseline & 73.47 & 56.00 & 53.80 & 10.02 & 28.29 & 1.93 & 61.09 & 13.41 \\
            FOCUS & 41.94 & 33.94 & 34.73 & 4.77 & 9.49 & 0.23 & 36.87 & 4.83 \\
            Token Distillation & 57.22 & 44.19 & 42.68 & 5.11 & 17.74 & 0.04 & 48.03 & 7.63 \\
            MATT & \textbf{66.39} & \textbf{48.50} & \textbf{47.11} & \textbf{5.20} & \textbf{20.68} & \textbf{0.90} & \textbf{54.00} & \textbf{8.93} \\
            \hline
        \end{tabular}
    }
\end{table*}

Table~\ref{table:multilingual-results} reports the means across the four target languages per model family; per-language detail is in Appendix~\ref{apx:multilingual-detail} (Table~\ref{table:multilingual-detail}). At the aggregate level, MATT outperforms FOCUS and Token Distillation on every metric in both family blocks, including both translation directions of Long FLORES and XL-Sum BLEU. Token Distillation slots between FOCUS and MATT, recovering most of FOCUS's discriminative gap and a smaller share of the generative gap, but does not match MATT's overall recovery in either family.

Looking at the per-language detail, MATT wins Avg Disc in 7 of the 8 (family $\times$ language) sub-blocks; the only exception is Qwen~3 4B on Japanese, where Token Distillation is marginally ahead (55.46 vs 54.46). MATT wins Avg Gen in 6 of the 8 sub-blocks; Token Distillation overtakes MATT on Avg Gen for Gemma~3 4B PT on Japanese and for Qwen~3 4B on German, in both cases primarily through stronger Long FLORES translation. The gap between MATT and the other methods is largest on Arabic and Swahili, where the base models' tokenizer coverage of the target language is poorest, and narrows on German and Japanese, where the base tokenizer is already comparatively efficient. A striking failure case is Qwen~3 4B on Swahili: Token Distillation degrades \emph{below} the FOCUS heuristic on Belebele (25.89 vs 33.78), suggesting that hidden-state matching is brittle when the teacher's representations for the target language are themselves weak.

A general observation about MATT in this setting is that it recovers what the teacher already knows, but cannot induce capabilities the teacher lacks; as a self-distillation method, its performance is upper-bounded by the original model. This is most visible on Qwen~3 4B for Swahili, where the base model itself is only marginally above random (Belebele 44.89, Global MMLU 38.25). MATT improves Belebele from FOCUS's 33.78 to 41.67 and Global MMLU from 26.50 to 33.00, narrowing the gap to the teacher but unable to surpass it. The same upper-bound argument applies to languages outside the teacher's pretraining mix entirely; cross-lingual transfer to such languages is beyond the scope of MATT.

\section{Conclusion}
\label{sec:conclusion}

In this work, we introduced MATT, a model-aware method for tokenizer transfer that leverages the internal dynamics of LLMs. We applied MATT to extend the tokenizers of Gemma 3 and Qwen 3 models across multiple languages and settings, demonstrating that it consistently recovers a large portion of the original model’s capabilities while requiring only a few GPU hours of training. Unlike heuristic-based methods that rely solely on the embedding layer, MATT refines token representations with direct feedback from the model, thanks to the novel Attention Influence Modeling (AIM) objective, allowing it to bridge the performance gap caused by tokenizer changes more effectively.

Our experiments highlight this advantage most clearly in the transfer of the 12 billion-parameter Gemma 3 model to Ukrainian. With the extended tokenizer introducing over 80,000 new tokens, MATT achieves an average score of 77.27 out of the original 78.18 on the discriminative tasks and 11.81 out of 12.68 on the generative ones, outperforming both heuristic and optimization-based baselines. This substantial improvement underscores the value of incorporating model dynamics into tokenizer transfer and shows that high performance can be retained at a fraction of the computational cost typically required for NTP training.

% \begin{ack}
% Use unnumbered first level headings for the acknowledgments. All acknowledgments

% Do {\bf not} include this section in the anonymized submission, only in the final paper. You can use the \texttt{ack} environment provided in the style file to automatically hide this section in the anonymized submission.
% \end{ack}

{
\small
\bibliographystyle{plainnat}
\bibliography{example_paper}
}

%%%%%%%%%%%%%%%%%%%%%%%%%%%%%%%%%%%%%%%%%%%%%%%%%%%%%%%%%%%%

\appendix

\section{Limitations}
\label{apx:limitations}
% The first limitation lies in the fact that MATT relies on tied input and output embeddings to fully realize its advantages. We outline possible strategies to relax this requirement in the Appendix~\ref{apx:untied-embeddings}.

% updated

The first limitation is that MATT only directly optimizes input embeddings. When input and output embeddings are tied, AIM's updates do propagate into the LM head, but the output embeddings still receive no \emph{direct} supervision toward producing good next-token distributions; this likely contributes to the residual gap on generative tasks observed throughout our experiments. For models with untied embeddings the issue is more pronounced, and we show in Appendix~\ref{apx:untied-embeddings} that adding an NTP loss to the AIM objective recovers most of that gap, at a slight cost in training efficiency.

Second, we do not perform continual pretraining with all weights unfrozen due to computational constraints, and instead evaluate only models with initialized or trained embedding layers. This is sufficient to compare MATT with existing baselines, whose primary goal is to provide a strong starting point for further adaptation.

A further limitation is the need for an additional forward pass during optimization through the model using the original tokenizer to obtain targets for the AIM objective. Although this adds computational overhead, the cost remains lower than a full forward pass through the entire model, as the target layer is positioned roughly one-third of the way through the network.

We have also not tested MATT on encoder-only architectures. In principle, applying it to such models would only require removing the causal constraint in the AIM definition.

Finally, we did not evaluate on truly low-resource languages, where benchmark coverage is too sparse for meaningful evaluation; Swahili is the closest proxy in our suite.

\section{Segmentation algorithm}
\label{apx:segmentation}

Instead of relying on word-based segmentation, we use an offset-based segmentation strategy. Designing a consistent word segmentation across tokenizers is challenging because tokenizers often differ in normalization rules, pre-tokenization steps, language coverage, etc. These differences make it difficult to ensure that segment boundaries match at the word level.

\begin{figure}[h]
    \centering
    \includegraphics[width=\linewidth]{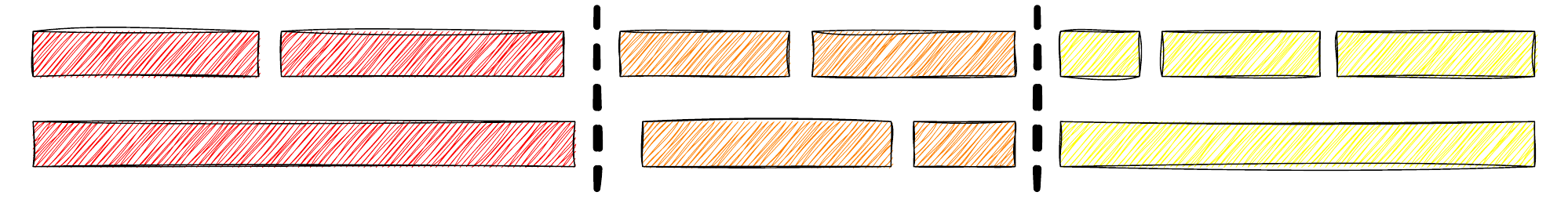}
    \caption{Offset-based segmentation algorithm visualization.}
    \label{fig:offset-based-segmentation}
\end{figure}

The offset-based method addresses this by operating directly on character offsets in the original text. Given two different tokenizations of the same string, along with the start and end positions of each token, the algorithm searches for all possible split positions that never cut through the middle of any token (see Figure~\ref{fig:offset-based-segmentation}).

This approach is universal: such a segmentation always exists, even if the worst case reduces to a single segment spanning the entire input. It can also lead to more precise alignments because the target tokenizer may break a word into several sub-tokens. By working with character offsets, we can introduce mid-word segment boundaries whenever they yield a better match.

For example, consider the sentence \textit{CH4 is a formula for methane}. Suppose the original tokenizer produces the tokens \texttt{\_for}, \texttt{m}, and \texttt{ula} for the word \textit{formula}, while the new tokenizer produces \texttt{\_form} and \texttt{ula}. A word-level strategy would force alignment at the whole-word boundary, but an offset-based method can instead match \texttt{\_for} and \texttt{m} with \texttt{\_form}, and \texttt{ula} with \texttt{ula}, which more closely respects both tokenizations.

Algorithm~\ref{alg:offset-based-segmentation} provides detailed pseudocode for implementation.

\begin{algorithm}[htbp]
  \caption{Offset–Based Segmentation}
  \label{alg:offset-based-segmentation}
  
  % Define the custom colored comment command
  \newcommand{\ColorComment}[1]{\COMMENT{\textcolor{purple}{#1}}}

  \begin{algorithmic}[1]
    \STATE {\bfseries Input:} teacher offsets $O_t$, student offsets $O_s$
    \STATE {\bfseries Output:} teacher segment ids $S_t$, student segment ids $S_s$
    
    \STATE \vspace{0.5em} \ColorComment{initialize outputs and counters}
    \STATE $S_t \gets [\,], S_s \gets [\,]$
    \STATE $e \gets -1$ \ColorComment{current end}
    \STATE $k \gets -1$ \ColorComment{current segment id}

    \STATE \vspace{0.5em} \ColorComment{iterate until both queues empty}
    \WHILE{$O_t \neq \emptyset$ \OR $O_s \neq \emptyset$}
      
      \STATE \vspace{0.3em} \ColorComment{if one side empty, label all remaining tokens with current segment id}
      \IF{$O_t = \emptyset$}
        \FOR{each $o$ in $O_s$}
          \STATE Append $k$ to $S_s$
        \ENDFOR
        \STATE \textbf{break}
      \ELSIF{$O_s = \emptyset$}
        \FOR{each $o$ in $O_t$}
          \STATE Append $k$ to $S_t$
        \ENDFOR
        \STATE \textbf{break}
      \ENDIF

      \STATE \vspace{0.3em} \ColorComment{peek next offsets}
      \STATE $(t_s, t_e) \gets \text{peek}(O_t)$
      \STATE $(s_s, s_e) \gets \text{peek}(O_s)$

      \STATE \vspace{0.3em} \ColorComment{continue with the same segment if overlap}
      \IF{$t_s < e$}
        \STATE Append $k$ to $S_t$, pop($O_t$)
        \STATE $e \gets \max(e, t_e)$
      \ELSIF{$s_s < e$}
        \STATE Append $k$ to $S_s$, pop($O_s$)
        \STATE $e \gets \max(e, s_e)$
      \ELSE
        \STATE \ColorComment{else start a new segment}
        \STATE $k \gets k + 1$
        \STATE Append $k$ to $S_t$ and $S_s$, pop($O_t$), pop($O_s$)
        \STATE $e \gets \max(t_e, s_e)$
      \ENDIF
    \ENDWHILE
    
    \STATE \vspace{0.5em} \textbf{return} $(S_t, S_s)$
  \end{algorithmic}
\end{algorithm}

\section{Convergence speed}
\label{apx:convergence-speed}

This appendix reports a convergence experiment with Gemma 3 4B PT trained on HPLT 2.0 Cleaned using a single multilingual extended tokenizer covering Arabic, German, English, Japanese, and Swahili; its compression statistics are reported in Table~\ref{table:convergence-tokenizer-stats}. We save model checkpoints every 3,000 training steps; the 250k-step row of Table~\ref{table:convergence-speed} corresponds to the fully trained model, and the earlier checkpoints allow us to observe how quickly the embeddings adapt to the new tokenizer and to evaluate whether training can be substantially shortened.

\begin{table}[h]
    \small
    \caption{Comparison of the original and the multilingual extended Gemma tokenizer used in this appendix. Compression rate is the average number of characters represented by a single token (higher is better).}
    \begin{center}
        \label{table:convergence-tokenizer-stats}
        \begin{tabular}{l|c|ccccc}
            \hline
            \textbf{Tokenizer}
            & \textbf{Vocabulary}
            & \multicolumn{5}{c}{\textbf{Compression Rate}} \\
            & \textbf{Size}
            & \textbf{ar} & \textbf{de} & \textbf{en} & \textbf{ja} & \textbf{sw} \\
            \hline
            original & 262{,}145 & 2.85 & 3.97 & 4.32 & 1.68 & 2.98 \\
            extended & 387{,}980 & \textbf{3.91} & \textbf{4.50} & \textbf{4.34} & \textbf{2.13} & \textbf{4.25} \\
            \hline
        \end{tabular}
    \end{center}
\end{table}

The AIM objective provides a rich learning signal for tuning the embeddings. Using a mean squared error (MSE) loss, the number of value pairs contributing to the objective is proportional to the product of the head dimensionality, the number of attention heads, the number of possible segment pairs, and the batch size. In the configuration used here -- four documents per batch, each truncated to 256 tokens -- this amounts to hundreds of millions of pairs at every step of the training.

\begin{figure}[h]
    \centering
        \begin{subfigure}{0.45\textwidth}
            \centering
            \includegraphics[width=\textwidth]{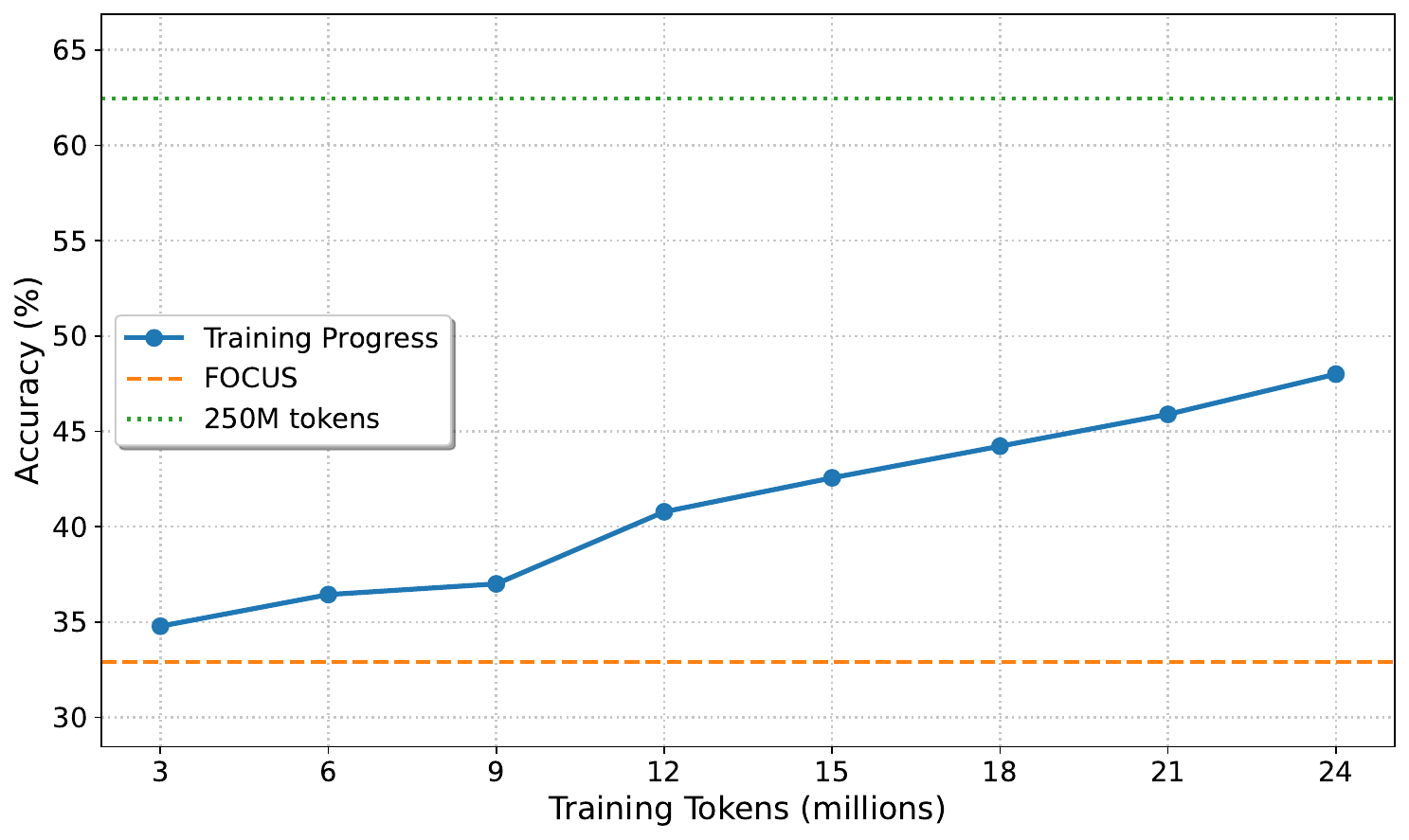}
            \caption{Arabic}
        \end{subfigure}
        \hfill
        \begin{subfigure}{0.45\textwidth}
            \centering
            \includegraphics[width=\textwidth]{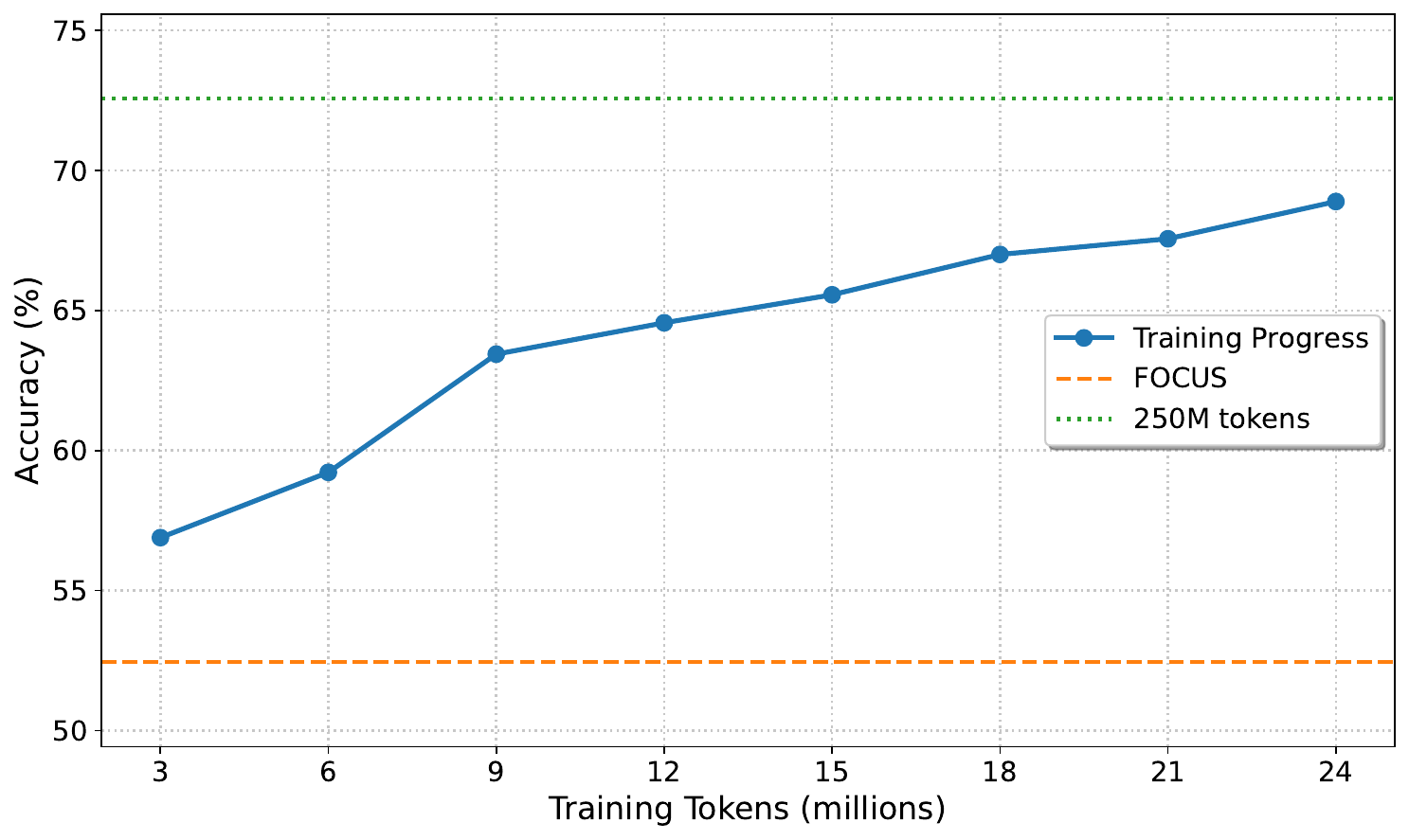} 
            \caption{German}
        \end{subfigure}
    
        \begin{subfigure}{0.45\textwidth}
            \centering
            \includegraphics[width=\textwidth]{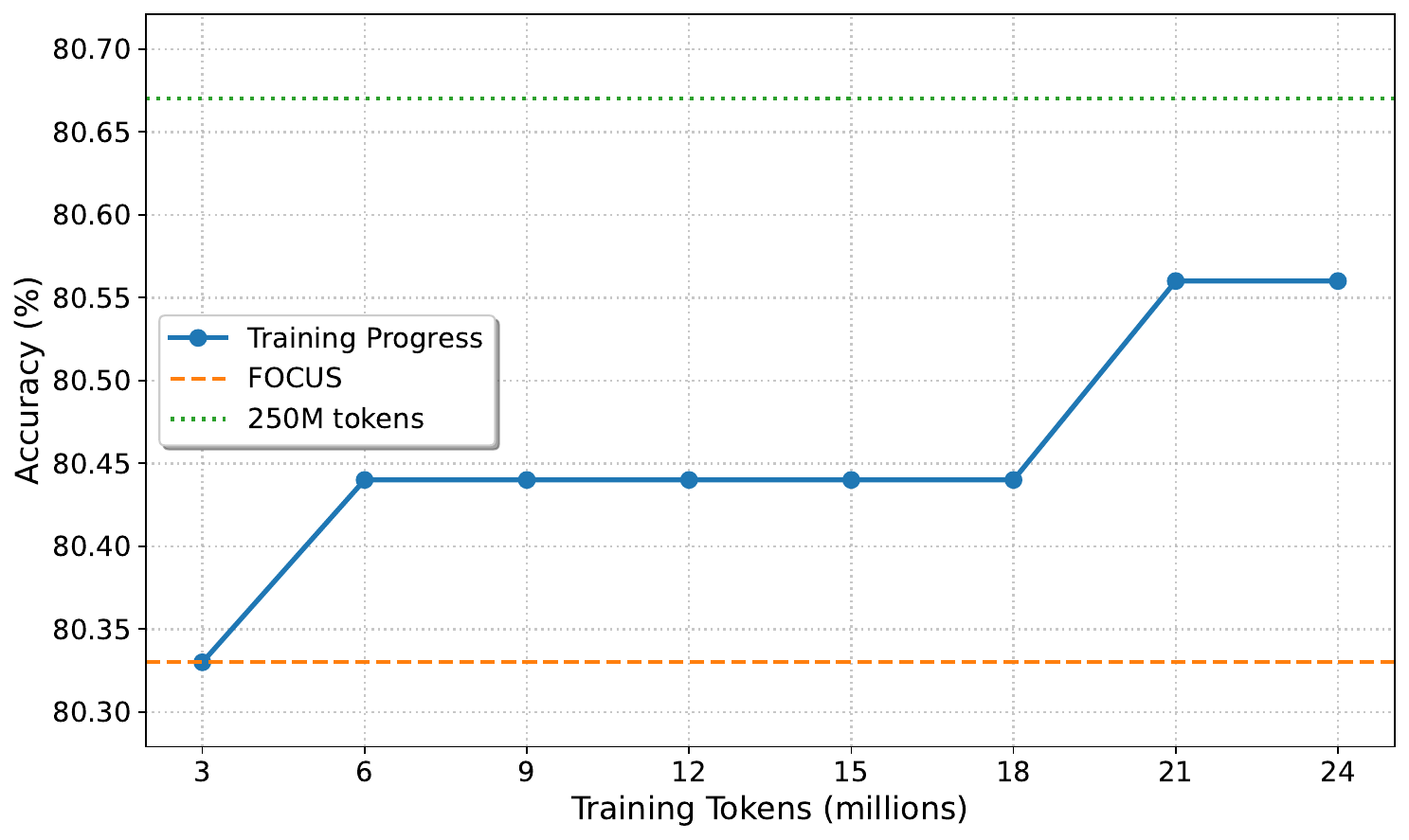}
            \caption{English}
        \end{subfigure}
        \hfill
        \begin{subfigure}{0.45\textwidth}
            \centering
            \includegraphics[width=\textwidth]{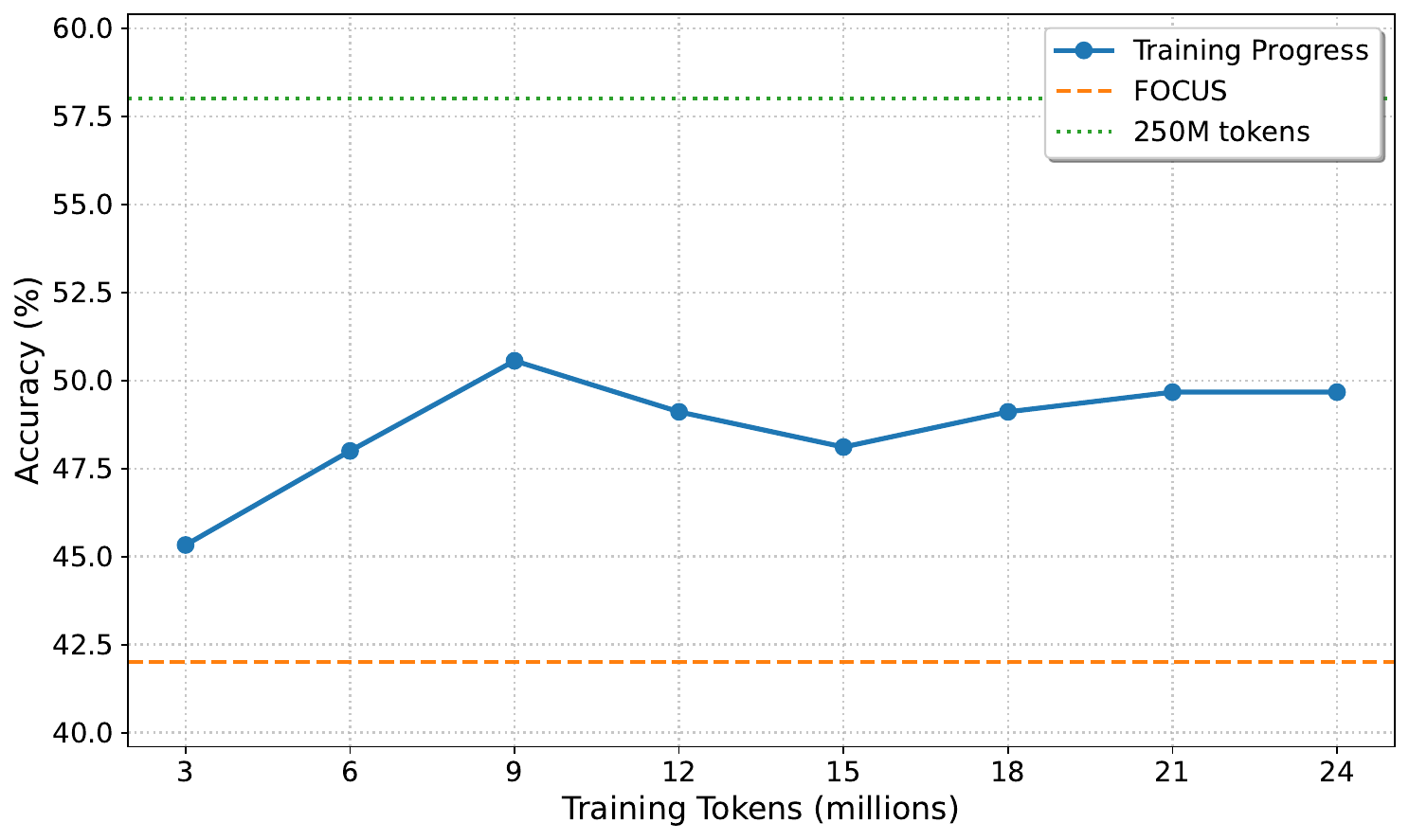}
            \caption{Japanese}
        \end{subfigure}
    
        \begin{subfigure}{0.45\textwidth}
            \centering
            \includegraphics[width=\textwidth]{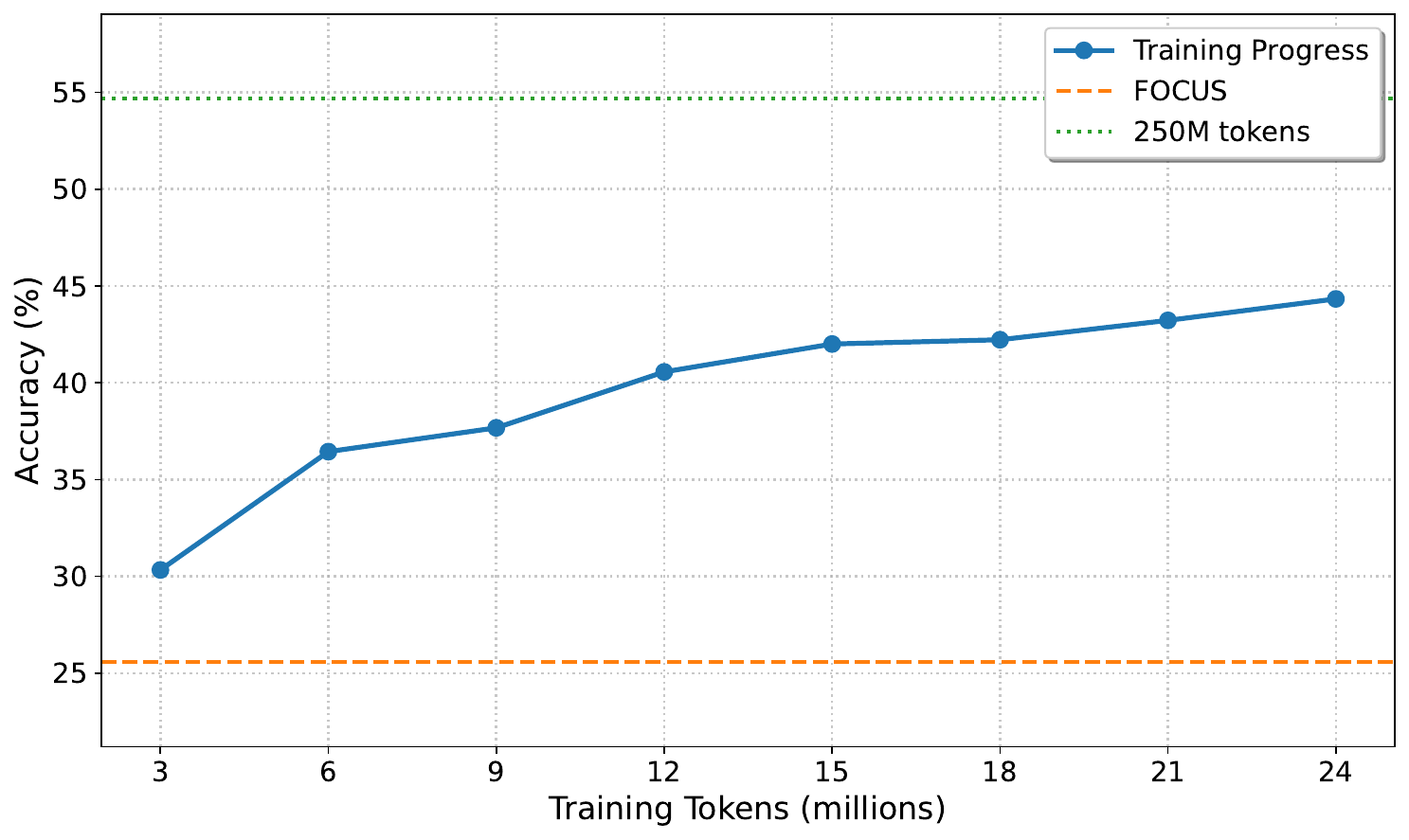}
            \caption{Swahili}
        \end{subfigure}
    
    \caption{Accuracy on the Belebele benchmark over training tokens for five languages, with horizontal lines marking FOCUS initialization and full performance.}
    \label{fig:convergence-speed}
\end{figure}

Table~\ref{table:convergence-speed} presents the results for the first eight checkpoints on the Belebele benchmark across all tested languages, while Figure~\ref{fig:convergence-speed} provides a visual view of the same trends.

The data show that more than 50\% of the final performance gains can be achieved in under 10\% of the total training steps, corresponding to fewer than five million tokens per language. This indicates that training time could be cut dramatically with only a minor loss in accuracy, especially if an adaptive data selection strategy is used to prioritize documents that contain a higher proportion of previously unseen tokens.

\renewcommand{\arraystretch}{1.2}
\begin{table}[h]
    \caption{Performance on the Belebele benchmark during early training of Gemma 3 4B PT with Model-Aware Tokenizer Transfer, showing rapid gains within the first 10\% of steps compared to the full run.}
    \begin{center}
        \label{table:convergence-speed} 
        \begin{tabular}{c|c|ccccc}                
            \hline
            & 
            & \multicolumn{5}{c}{\textbf{Belebele}} \\
            \textbf{Steps \#}
            & \textbf{Tokens \#}
            & \textbf{ar} & \textbf{de} & \textbf{en} & \textbf{ja} & \textbf{sw} \\
            \hline
            0k & 0M & 32.89 & 52.44 & 80.33 & 42.00 & 25.56 \\
            \hdashline
            3k & 3M & 34.78 & 56.89 & 80.33 & 45.33 & 30.33 \\
            6k & 6M & 36.44 & 59.22 & 80.44 & 48.00 & 36.44 \\
            9k & 9M & 37.00 & 63.44 & 80.44 & 50.56 & 37.67 \\
            12k & 12M & 40.78 & 64.56 & 80.44 & 49.11 & 40.56 \\
            15k & 15M & 42.56 & 65.56 & 80.44 & 48.11 & 42.00 \\
            18k & 18M & 44.22 & 67.00 & 80.44 & 49.11 & 42.22 \\
            21k & 21M & 45.89 & 67.56 & 80.56 & 49.67 & 43.22 \\
            24k & 24M & 48.00 & 68.89 & 80.56 & 49.67 & 44.33 \\
            \hdashline
            250k & 250M & 62.44 & 72.56 & 80.67 & 58.00 & 54.67
            \\
            \hline
        \end{tabular}
        
    \end{center}
\end{table}

% \section{More Results}
% \label{apx:more-results}
% \input{latex/d-more-results}

\section{Ablation studies}
\label{apx:ablation-studies}

We apply the MATT method with AIM objective, transferring Gemma 3 4B PT to a Ukrainian-centric tokenizer by \citet{zaduha2025post9194}, which increases the compression rate by around 50\%. We conduct several ablation studies to determine the optimal training configuration, evaluating performance on the Belebele and Global MMLU benchmarks in the same setting as in Section~\ref{sec:experiments}. We report the results in Table~\ref{table:ablation-studies} and describe our insights below.

\renewcommand{\arraystretch}{1.1}
\begin{table}[h]
    \caption{Ablation studies of MATT configurations. \textbf{3} and \textbf{5} denote the number of layers used for AIM objective.}
    
    \begin{center}
        \label{table:ablation-studies} 
        \begin{tabular}{l||cc;{2pt/2pt}cc||cc;{2pt/2pt}cc}
            
            \hline
                & \multicolumn{2}{c;{2pt/2pt}}{\textbf{VRAM (GiB)}}
                & \multicolumn{2}{c||}{\textbf{Time}}
                & \multicolumn{2}{c;{2pt/2pt}}{\textbf{Belebele}}
                & \multicolumn{2}{c}{\textbf{Global MMLU}}
                \\
                & \textbf{3}
                & \textbf{5}
                & \textbf{3}
                & \textbf{5}
                & \textbf{3}
                & \textbf{5}
                & \textbf{3}
                & \textbf{5}
                \\
            \hline
                \multicolumn{9}{c}{\textbf{All Layers vs. Last Layer}} \\
            \hline
                all layers & 17.3 & 26.5 & 1h 33m & 2h 19m & 32.56 & 35.11 & 28.63 & 29.00 \\
                last layer & \textbf{9.1} & \textbf{10.1} & \textbf{0h 56m} & \textbf{1h 04m} & \textbf{37.22} & \textbf{60.11} & \textbf{29.95} & \textbf{34.80} \\
            \hline
                \multicolumn{9}{c}{\textbf{Initialization Method}} \\
            \hline
                WECHSEL & \textbf{9.1} & \textbf{10.1} & \textbf{0h 52m} & \textbf{1h 04m} & 42.22 & 59.78 & 29.82 & 33.56 \\
                FOCUS & \textbf{9.1} & \textbf{10.1} & 0h 55m & 1h 06m & 37.11 & \textbf{60.89} & 30.10 & 34.72 \\
                Transtokenizers & \textbf{9.1} & \textbf{10.1} & 0h 55m & \textbf{1h 04m} & \textbf{52.44} & \textbf{60.89} & \textbf{32.43} & \textbf{34.75} \\
            \hline
        \end{tabular}
        
    \end{center}
\end{table}

\paragraph{AIM objective on all layers deteriorates both efficiency and performance compared to only the last one.} 
Since AIM is defined for a single attention layer, we can apply it many times to any subset of layers.  To balance efficiency and accuracy, we compare defining the AIM objective over all layers up to a target depth $n$ against using only the final $n$-th layer. Results in Table~\ref{table:ablation-studies} show that restricting AIM to the last layer requires much less VRAM and training time, while also delivering better downstream performance.

\paragraph{FOCUS and Transtokenizers perform similarly on higher layers, while WECHSEL underperforms.} Because MATT is independent of the embedding initialization method, different starting points can be tested. We compare WECHSEL, FOCUS, and Transtokenizers. FOCUS and Transtokenizers perform similarly on higher layers, while WECHSEL lags behind (see Table~\ref{table:ablation-studies}). Although Transtokenizers occasionally achieves the best scores, in other experiments, we find FOCUS to be more stable across models, and therefore make it our default choice. 

Transtokenizers method may have an upper hand due to its better utilization of English embeddings, as it learns an English-Ukrainian token-level dictionary from parallel corpora and utilizes it to transfer embeddings from English tokens to their Ukrainian counterparts. Whereas FOCUS utilizes the tokens' overlap to train a FastText model over it, and although it contains English tokens as well, the FastText training corpus contains little data that encompasses both English and Ukrainian text in the same document. This could potentially limit the FOCUS to pay attention mostly to Ukrainian overlapped tokens, given the limited usability of English tokens’ embeddings.

MATT, in the way it uses existing embeddings, is conceptually closer to FOCUS than Transtokenizers, as it models inter-token communication of original tokens, focusing predominantly on Ukrainian ones. As denoted in Appendix~\ref{apx:convergence-speed}, MATT quickly converges, requiring little data to recover a large part of the original model’s performance. This means that small differences in the initialization (the difference between average scores for FOCUS and Transtokenizers is less than 10\% in the case of Gemma 3 12B PT; see Table~\ref{table:main-results}) are evened out during training. This can also be seen with WECHSEL, which performs considerably worse compared to FOCUS or Transtokenizers (Table~\ref{table:main-results}), but achieves only slightly worse results after a round of MATT training, especially on higher layers (see Table~\ref{table:ablation-studies}). The final observation is presented in Table~\ref{table:main-results}, where we see that even the NTP baseline starting from the Transtokenizers initialization, which is initially better, achieves similar performance to the NTP over FOCUS.

Minor efficiency differences in Table~\ref{table:ablation-studies} are likely due to external factors such as checkpointing overhead.

\begin{figure}[h]
    \centering
    \begin{subfigure}{0.47\textwidth}
        \centering
        \includegraphics[width=\linewidth]{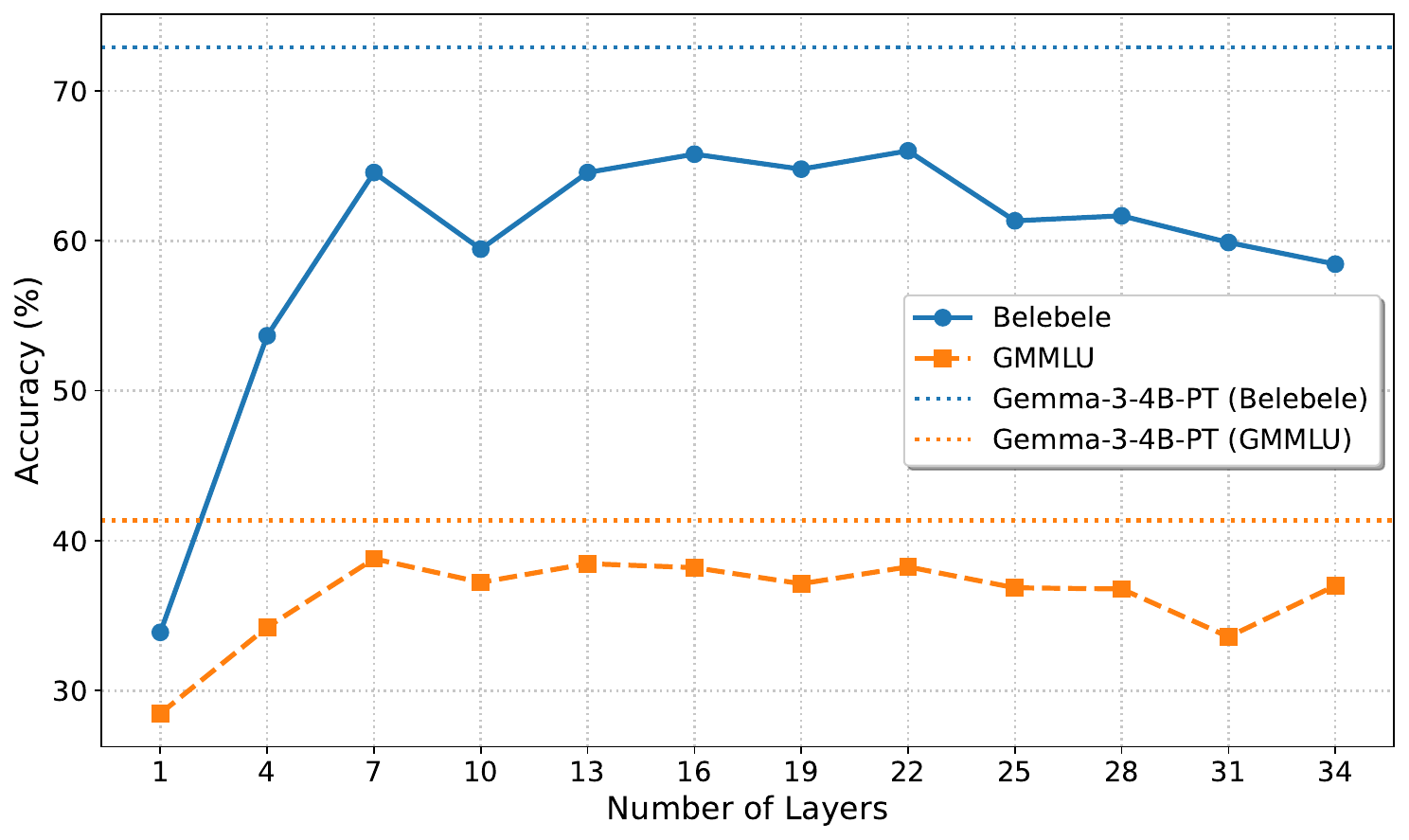}
        \caption{Performance}
        \label{fig:ablation_n_layers_a}
    \end{subfigure}%
    \hfill
    \begin{subfigure}{0.47\textwidth}
        \centering
        \includegraphics[width=\linewidth]{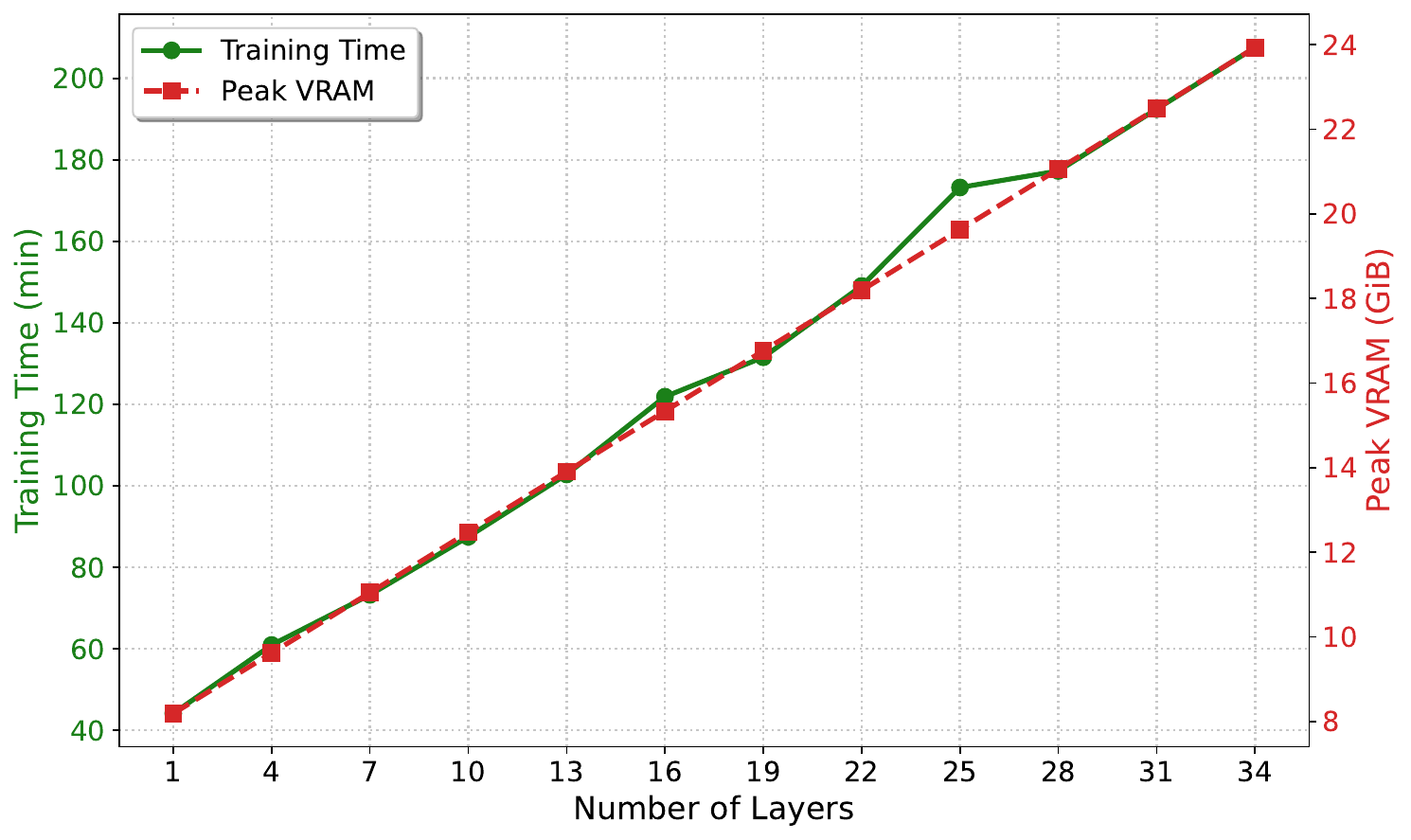}
        \caption{Efficiency}
        \label{fig:ablation_n_layers_b}
    \end{subfigure}
    
    \caption{Effect of applying the AIM objective to different numbers of layers. The plots show the trade-off between model performance (a) and computational efficiency (b) as the application depth increases.}
    \label{fig:ablation_n_layers}
\end{figure}

\paragraph{AIM on higher layers leads to better results, but saturates at around one third of the model's depth.} 
MATT allows selecting how deep into the model the AIM objective is applied, creating a natural trade-off between efficiency and accuracy. We evaluate different target depths and find that performance steadily improves as AIM is applied to higher layers (see Figure~\ref{fig:ablation_n_layers}), but gains saturate once the objective reaches roughly one third of the model’s total depth. In contrast, memory consumption and training time continue to grow almost linearly with the number of layers, highlighting the cost of deeper alignment.

\section{Performance on models with tied embeddings}
\label{apx:untied-embeddings}
The use of tied embeddings varies greatly both between the model families and model sizes. For example, Llama 3.2 1B  and Llama 3.2 3B \citep{grattafiori2024llama} both utilize tied embeddings to reduce the number of parameters, whereas a larger Llama 3.1 8B does not. In contrast, the Gemma family \citep{team2024gemma, team2024gemma2, team2025gemma} consistently uses tied embeddings across all sizes.

A significant amount of model pretraining research conducts very little to no experimentation on the effects of embedding tying \citep{jiang2023mistral7b, touvron2023llama, groeneveld2024olmo, qwen3technicalreport}. And those that do \citep{bai2023qwen} offer a limited explanation of the reasons behind their choice, referring to preliminary results that are not reported in the paper.

More recent research suggests that embedding tying is more effective both from a theoretical standpoint \citep{bertolotti2024by} and in achieving lower validation loss and better performance on downstream tasks \citep{allal2025_the_smol_training_playbook_the_secrets_to_building_world_class_llms}. This leads us to believe that the share of models with tied embeddings may increase in the coming years, making our method even more relevant.

We conducted additional experiments on Qwen~3 8B~\cite{qwen3technicalreport}, which does not tie embeddings. The presented experiment uses a single multilingual extended Qwen tokenizer covering Arabic, German, English, Japanese, and Swahili (i.e., one tokenizer trained on a multilingual mixture rather than the per-language tokenizers used in Section~\ref{subsec:multilingual-results}); its compression statistics are reported in Table~\ref{table:untied-tokenizer-stats}.

\begin{table}[h]
    \small
    \caption{Comparison of the original and the multilingual extended Qwen tokenizer used in this appendix. Compression rate is the average number of characters represented by a single token (higher is better).}
    \begin{center}
        \label{table:untied-tokenizer-stats}
        \begin{tabular}{l|c|ccccc}
            \hline
            \textbf{Tokenizer}
            & \textbf{Vocabulary}
            & \multicolumn{5}{c}{\textbf{Compression Rate}} \\
            & \textbf{Size}
            & \textbf{ar} & \textbf{de} & \textbf{en} & \textbf{ja} & \textbf{sw} \\
            \hline
            original & 151{,}669 & 2.60 & 3.47 & 4.36 & 1.49 & 2.58 \\
            extended & 298{,}833 & \textbf{3.92} & \textbf{4.49} & \textbf{4.42} & \textbf{2.29} & \textbf{4.23} \\
            \hline
        \end{tabular}
    \end{center}
\end{table} We train on a limited amount of data relative to the original experimental setting due to resource constraints. The AIM objective used in MATT is enhanced by incorporating an NTP loss, similar to \citet{dobler2026token}, to appropriately handle output embeddings in a model with untied embeddings. The losses are aggregated by aligning their scales, rather than by using the norm of the gradients \cite{pmlr-v80-chen18a}, which leaves room for further improvement. We apply AIM on the 16th layer out of 36.

\begin{table}[h]
    \caption{Benchmark results for transferring original tokenizers to their extended versions across five languages (Arabic, German, English, Japanese, Swahili) for Qwen~3 8B with untied embeddings.}
    \resizebox{\linewidth}{!}{%
            \centering
            \label{table:untied-embeddings-results} 
            \begin{tabular}{l|c|ccccc|ccccc|ccccc|c}
                
                \hline
                    % & \textbf{VRAM}
                    & \textbf{Time}
                    & \multicolumn{5}{c|}{\textbf{Belebele}}
                    & \multicolumn{5}{c|}{\textbf{MMMLU}}
                    & \multicolumn{5}{c|}{\textbf{Global MMLU}}
                    & \textbf{Avg}
                    \\
                    \textbf{Model}
                    % & \textbf{GiB}
                    &
                    
                    & \textbf{ar}
                    & \textbf{de}
                    & \textbf{en}
                    & \textbf{ja}
                    & \textbf{sw}
    
                    & \textbf{ar}
                    & \textbf{de}
                    & \textbf{en}
                    & \textbf{ja}
                    & \textbf{sw}
    
                    & \textbf{ar}
                    & \textbf{de}
                    & \textbf{en}
                    & \textbf{ja}
                    & \textbf{sw}
                    \\

                % qwen
                \hline
                Qwen3 8B
                & - %& -
                & 86.11 & 91.56 & 92.56 & 84.00 & 60.22
                & 61.91 & 68.25 & 74.56 & 65.77 & 41.20
                & 66.25 & 71.25 & 77.75 & 66.50 & 45.25
                & 70.21
                \\
                \hdashline
                FOCUS
                & - %& -
                & 37.33 & 60.44 & 90.78 & 41.67 & 29.22
                & 32.62 & 43.90 & 70.78 & 37.16 & 30.07
                & 30.75 & 45.00 & \textbf{72.50} & 34.25 & 30.50
                & 45.80
                \\
                FOCUS w/ NTP
                & 4h 10m %& -
                & 57.44 & 79.89 & 89.67 & 51.11 & 35.00
                & 38.35 & 47.57 & 68.60 & 38.95 & 31.53
                & 38.25 & 51.25 & \textbf{72.50} & 40.25 & 29.75
                & 51.34
                \\
                MATT w/ NTP
                & 1h 40m %& -
                & 50.78 & 63.33 & 84.33 & 38.00 & 30.00
                & 45.43 & 57.38 & 72.25 & 45.19 & 32.46
                & 45.25 & 56.75 & \textbf{72.50} & 41.00 & \textbf{32.00}
                & 51.11
                \\
                MATT w/ NTP
                & \textbf{3h 20m} %& 3h 38m
                & \textbf{77.78} & \textbf{85.44} & \textbf{91.67} & \textbf{64.33} & \textbf{38.00}
                & \textbf{48.56} & \textbf{59.36} & \textbf{72.31} & \textbf{46.71} & \textbf{33.18}
                & \textbf{50.75} & \textbf{58.25} & 71.75 & \textbf{45.50} & 31.50
                & \textbf{58.34}
                \\
                \hline
            \end{tabular}
            
    }
\end{table}

MATT, with NTP loss and unfrozen embeddings, surpasses both the FOCUS and NTP baselines by a considerable margin. It achieved the performance of the NTP baseline with approximately 40\% of the compute budget allocated to it, underlining the importance of a rich signal coming from the AIM objective. Further training only improved the results, widening the gap between MATT and baselines.

Although training is slower than with tied embeddings, MATT still outperforms both heuristic and optimization-based baselines even in this setting. Considering that the training runs were shorter than those presented in Section~\ref{subsec:multilingual-results}, we can expect further growth in performance.

\clearpage
\section{Per-language multilingual results}
\label{apx:multilingual-detail}
Per-language details for the multilingual setting in Section~\ref{subsec:multilingual-results}. Table~\ref{table:multilingual-tokenizer-stats} reports the per-language extended tokenizer statistics: original and extended vocabulary sizes, and the compression rate (average number of characters per token) measured on a held-out HPLT 2.0 sample for each target language. Table~\ref{table:multilingual-detail} reports the per-language adaptation results aggregated to means in Table~\ref{table:multilingual-results}; training-time wall-clocks (single H100) appear in the Time column, where Token Distillation runs were configured to match or marginally exceed MATT's per-language compute budget. XL-Sum has no German subset (denoted ``--'').

\begin{table}[h]
    \small
    \caption{Per-language extended tokenizer statistics. Compression rate is the average number of characters represented by a single token (higher is better), measured on a held-out HPLT 2.0 sample of the target language.}
    \begin{center}
        \label{table:multilingual-tokenizer-stats}
        \begin{tabular}{ll|cc|cc}
            \hline
            & & \multicolumn{2}{c|}{\textbf{Original}} & \multicolumn{2}{c}{\textbf{Extended}} \\
            \textbf{Family} & \textbf{Lang}
            & \textbf{Vocab} & \textbf{Compr.}
            & \textbf{Vocab} & \textbf{Compr.}
            \\
            \hline
            Gemma 3 (4B / 12B PT) & ar & 262{,}145 & 2.85 & 344{,}152 & \textbf{4.15} \\
                                  & de & 262{,}145 & 3.97 & 325{,}863 & \textbf{4.65} \\
                                  & ja & 262{,}145 & 1.68 & 338{,}663 & \textbf{2.22} \\
                                  & sw & 262{,}145 & 2.98 & 331{,}378 & \textbf{3.82} \\
            \hline
            Qwen 3 (0.6B / 4B / 8B) & ar & 151{,}669 & 2.60 & 241{,}160 & \textbf{4.19} \\
                                    & de & 151{,}669 & 3.47 & 229{,}814 & \textbf{4.58} \\
                                    & ja & 151{,}669 & 1.49 & 239{,}381 & \textbf{2.41} \\
                                    & sw & 151{,}669 & 2.58 & 230{,}444 & \textbf{3.54} \\
            \hline
        \end{tabular}
    \end{center}
\end{table}

\renewcommand{\arraystretch}{1.2}
\begin{table*}[!htbp]
    \caption{Per-language multilingual results for Gemma 3 4B PT and Qwen 3 4B with their tokenizers extended per language, evaluated on the target language only. Training time is wall-clock on a single H100.}
    \resizebox{\linewidth}{!}{%
        \centering
        \label{table:multilingual-detail}
        \begin{tabular}{l|c|cccccc|cc}
            \hline
                \textbf{Method}
                & \begin{tabular}{@{}c@{}}\textbf{Training} \\ \textbf{Time}\end{tabular}
                & \textbf{Belebele}
                & \begin{tabular}{@{}c@{}}\textbf{Global} \\ \textbf{MMLU}\end{tabular}
                & \textbf{MMMLU}
                & \multicolumn{2}{c}{\begin{tabular}{@{}cc@{}}\multicolumn{2}{c}{\textbf{Long FLORES}} \\ \textbf{en$\rightarrow$x} & \textbf{x$\rightarrow$en}\end{tabular}}
                & \textbf{XL-Sum}
                & \begin{tabular}{@{}c@{}}\textbf{Avg} \\ \textbf{Disc}\end{tabular}
                & \begin{tabular}{@{}c@{}}\textbf{Avg} \\ \textbf{Gen}\end{tabular}
            \\
            \hline
            \multicolumn{10}{c}{\textbf{Gemma 3 4B PT}} \\
            \hdashline
            \multicolumn{10}{c}{\textit{Arabic}} \\
            \hdashline
            Baseline & -- & 69.67 & 55.25 & 48.74 & 5.07 & 20.21 & 5.12 & 57.89 & 10.13 \\
            FOCUS & -- & 36.44 & 33.75 & 31.61 & 0.14 & 4.60 & 0.21 & 33.94 & 1.65 \\
            Token Distillation & 7h 29m & 52.11 & 42.25 & 41.29 & 0.01 & 11.40 & \textbf{1.42} & 45.22 & 4.28 \\
            MATT & 5h 02m & \textbf{66.67} & \textbf{51.00} & \textbf{45.68} & \textbf{0.88} & \textbf{12.87} & 0.92 & \textbf{54.45} & \textbf{4.89} \\
            \hdashline
            \multicolumn{10}{c}{\textit{German}} \\
            \hdashline
            Baseline & -- & 75.67 & 60.75 & 53.55 & 15.55 & 24.46 & -- & 63.32 & 20.01 \\
            FOCUS & -- & 60.67 & 46.50 & 42.75 & 6.25 & 16.29 & -- & 49.97 & 11.27 \\
            Token Distillation & 6h 48m & 66.78 & 57.00 & 48.64 & 4.95 & \textbf{20.49} & -- & 57.47 & 12.72 \\
            MATT & 5h 39m & \textbf{73.56} & \textbf{59.25} & \textbf{52.35} & \textbf{9.26} & 18.29 & -- & \textbf{61.72} & \textbf{13.77} \\
            \hdashline
            \multicolumn{10}{c}{\textit{Japanese}} \\
            \hdashline
            Baseline & -- & 69.89 & 53.75 & 50.39 & 0.20 & 10.97 & 0.39 & 58.01 & 3.85 \\
            FOCUS & -- & 44.56 & 42.00 & 38.76 & 0.00 & 5.66 & 0.02 & 41.77 & 1.89 \\
            Token Distillation & 6h 31m & 59.56 & 42.75 & 44.21 & \textbf{0.02} & \textbf{7.63} & \textbf{0.08} & 48.84 & \textbf{2.58} \\
            MATT & 5h 21m & \textbf{64.33} & \textbf{53.00} & \textbf{46.41} & 0.01 & 6.49 & 0.01 & \textbf{54.58} & 2.17 \\
            \hdashline
            \multicolumn{10}{c}{\textit{Swahili}} \\
            \hdashline
            Baseline & -- & 66.89 & 50.25 & 44.35 & 10.88 & 25.23 & 6.90 & 53.83 & 14.34 \\
            FOCUS & -- & 31.67 & 39.75 & 33.39 & 1.33 & 7.01 & 0.32 & 34.94 & 2.88 \\
            Token Distillation & 8h 07m & 51.11 & 41.25 & 39.93 & 0.26 & 11.95 & 1.76 & 44.10 & 4.66 \\
            MATT & 5h 21m & \textbf{59.56} & \textbf{48.00} & \textbf{42.92} & \textbf{5.63} & \textbf{15.24} & \textbf{2.88} & \textbf{50.16} & \textbf{7.92} \\
            \hline
            \multicolumn{10}{c}{\textbf{Qwen 3 4B}} \\
            \hdashline
            \multicolumn{10}{c}{\textit{Arabic}} \\
            \hdashline
            Baseline & -- & 82.44 & 58.00 & 56.15 & 9.95 & 35.23 & 3.36 & 65.53 & 16.18 \\
            FOCUS & -- & 38.33 & 28.50 & 31.81 & 2.62 & 8.61 & 0.40 & 32.88 & 3.88 \\
            Token Distillation & 7h 12m & 60.67 & 41.75 & 41.63 & 0.07 & 21.61 & 0.12 & 48.01 & 7.27 \\
            MATT & 6h 35m & \textbf{75.22} & \textbf{49.25} & \textbf{49.26} & \textbf{3.07} & \textbf{25.48} & \textbf{2.26} & \textbf{57.91} & \textbf{10.27} \\
            \hdashline
            \multicolumn{10}{c}{\textit{German}} \\
            \hdashline
            Baseline & -- & 86.22 & 66.00 & 62.86 & 25.83 & 40.86 & -- & 71.69 & 33.34 \\
            FOCUS & -- & 55.22 & 45.00 & 40.87 & 14.26 & 18.65 & -- & 47.03 & 16.45 \\
            Token Distillation & 7h 35m & 74.44 & 56.75 & 52.08 & \textbf{20.29} & \textbf{33.05} & -- & 61.09 & \textbf{26.67} \\
            MATT & 7h 45m & \textbf{82.78} & \textbf{61.50} & \textbf{57.93} & 15.30 & 29.99 & -- & \textbf{67.40} & 22.65 \\
            \hdashline
            \multicolumn{10}{c}{\textit{Japanese}} \\
            \hdashline
            Baseline & -- & 80.33 & 61.75 & 59.45 & 1.20 & 24.45 & 0.37 & 67.18 & 8.67 \\
            FOCUS & -- & 40.44 & 35.75 & 35.09 & \textbf{0.21} & 7.22 & 0.02 & 37.09 & 2.49 \\
            Token Distillation & 7h 35m & \textbf{67.89} & 49.50 & \textbf{49.00} & 0.00 & 15.96 & 0.00 & \textbf{55.46} & 5.32 \\
            MATT & 6h 30m & 65.89 & \textbf{50.25} & 47.24 & 0.11 & \textbf{17.36} & \textbf{0.10} & 54.46 & \textbf{5.86} \\
            \hdashline
            \multicolumn{10}{c}{\textit{Swahili}} \\
            \hdashline
            Baseline & -- & 44.89 & 38.25 & 36.72 & 3.10 & 12.60 & 2.05 & 39.95 & 5.91 \\
            FOCUS & -- & 33.78 & 26.50 & 31.14 & 1.98 & 3.47 & 0.25 & 30.47 & 1.90 \\
            Token Distillation & 8h 52m & 25.89 & 28.75 & 28.01 & 0.09 & 0.35 & 0.00 & 27.55 & 0.15 \\
            MATT & 7h 39m & \textbf{41.67} & \textbf{33.00} & \textbf{34.01} & \textbf{2.34} & \textbf{9.88} & \textbf{0.35} & \textbf{36.23} & \textbf{4.19} \\
            \hline
        \end{tabular}
    }
\end{table*}

A note on the Japanese generative metrics: the original Gemma 3 4B PT and Qwen 3 4B already score under 1.5 BLEU on Long FLORES en$\rightarrow$ja (0.20 and 1.20) and below 0.5 BLEU on XL-Sum (0.39 and 0.37), so cell-level differences between FOCUS, Token Distillation, and MATT in these columns sit within BLEU noise on a near-zero scale rather than reflecting meaningful differences in generation quality. The Long FLORES x$\rightarrow$en direction, where the model generates English, is comparatively well-behaved (10.97 and 24.45 BLEU on the Gemma and Qwen baselines respectively), and the same caution in milder form applies to other low-baseline cells, e.g.\ XL-Sum on Swahili for Qwen 3 4B (2.05).

\section{Licenses of used assets}
\label{apx:licenses}
All assets used in this work are publicly released by their original authors and used in accordance with their stated licenses. Table~\ref{table:asset-licenses} lists each asset together with its license and the URL where the license is declared, so that the reader can verify both the license and the version used.

\begin{table}[h]
    \caption{Licenses of assets used in this work. Links point to the page where each license is stated.}
    \label{table:asset-licenses}
    \centering
    \small
    \begin{tabular}{lll}
        \toprule
        \textbf{Asset} & \textbf{License} & \textbf{Source} \\
        \midrule
        \multicolumn{3}{l}{\textit{Base models}} \\
        Gemma 3 (4B / 12B PT) & Gemma Terms of Use & \href{https://ai.google.dev/gemma/terms}{ai.google.dev/gemma/terms} \\
        Qwen 3 (4B / 8B) & Apache 2.0 & \href{https://huggingface.co/Qwen/Qwen3-4B}{huggingface.co/Qwen/Qwen3-4B} \\
        \midrule
        \multicolumn{3}{l}{\textit{Training data}} \\
        Kobza & CC0 1.0 & \href{https://huggingface.co/datasets/Goader/kobza}{huggingface.co/datasets/Goader/kobza} \\
        HPLT 2.0 Cleaned & CC0 1.0 & \href{https://huggingface.co/datasets/HPLT/HPLT2.0_cleaned}{huggingface.co/datasets/HPLT/HPLT2.0\_cleaned} \\
        OpenSubtitles (OPUS) & Unspecified (research use) & \href{https://opus.nlpl.eu/OpenSubtitles.php}{opus.nlpl.eu/OpenSubtitles.php} \\
        NLLB & CC-BY-SA 4.0 & \href{https://huggingface.co/datasets/openlanguagedata/flores_plus}{huggingface.co/datasets/openlanguagedata/flores\_plus} \\
        \midrule
        \multicolumn{3}{l}{\textit{Evaluation benchmarks}} \\
        Belebele & CC-BY-SA 4.0 & \href{https://huggingface.co/datasets/facebook/belebele}{huggingface.co/datasets/facebook/belebele} \\
        Global MMLU & Apache 2.0 & \href{https://huggingface.co/datasets/CohereForAI/Global-MMLU}{huggingface.co/datasets/CohereForAI/Global-MMLU} \\
        MMMLU (OpenAI) & MIT & \href{https://huggingface.co/datasets/openai/MMMLU}{huggingface.co/datasets/openai/MMMLU} \\
        FLORES / Long FLORES & CC-BY-SA 4.0 & \href{https://huggingface.co/datasets/openlanguagedata/flores_plus}{huggingface.co/datasets/openlanguagedata/flores\_plus} \\
        WMT24++ & Apache 2.0 & \href{https://huggingface.co/datasets/google/wmt24pp}{huggingface.co/datasets/google/wmt24pp} \\
        XL-Sum & CC-BY-NC-SA 4.0 & \href{https://huggingface.co/datasets/csebuetnlp/xlsum}{huggingface.co/datasets/csebuetnlp/xlsum} \\
        \midrule
        \multicolumn{3}{l}{\textit{Frameworks and baseline implementations}} \\
        \texttt{lm-evaluation-harness} & MIT & \href{https://github.com/EleutherAI/lm-evaluation-harness}{github.com/EleutherAI/lm-evaluation-harness} \\
        WECHSEL & MIT & \href{https://github.com/CPJKU/wechsel}{github.com/CPJKU/wechsel} \\
        \texttt{transtokenizers} & MIT & \href{https://github.com/LAGoM-NLP/transtokenizer}{github.com/LAGoM-NLP/transtokenizer} \\
        TokAlign & MIT & \href{https://github.com/ZNLP/TokAlign}{github.com/ZNLP/TokAlign} \\
        FOCUS / \texttt{deepfocus} & MIT & \href{https://github.com/konstantinjdobler/focus}{github.com/konstantinjdobler/focus} \\
        \texttt{token-distillation} & MIT & \href{https://github.com/konstantinjdobler/token-distillation}{github.com/konstantinjdobler/token-distillation} \\
        \bottomrule
    \end{tabular}
\end{table}

\clearpage
\section{Token-level AIM visualization}
\label{apx:aim-visualization}
This appendix accompanies Section~\ref{sec:method}; it complements Figure~\ref{fig:aim} with a sequence-level view of the same alignment. Notation matches Section~\ref{sec:method}.

\begin{figure}[h]
    \centering
    \includegraphics[width=\linewidth]{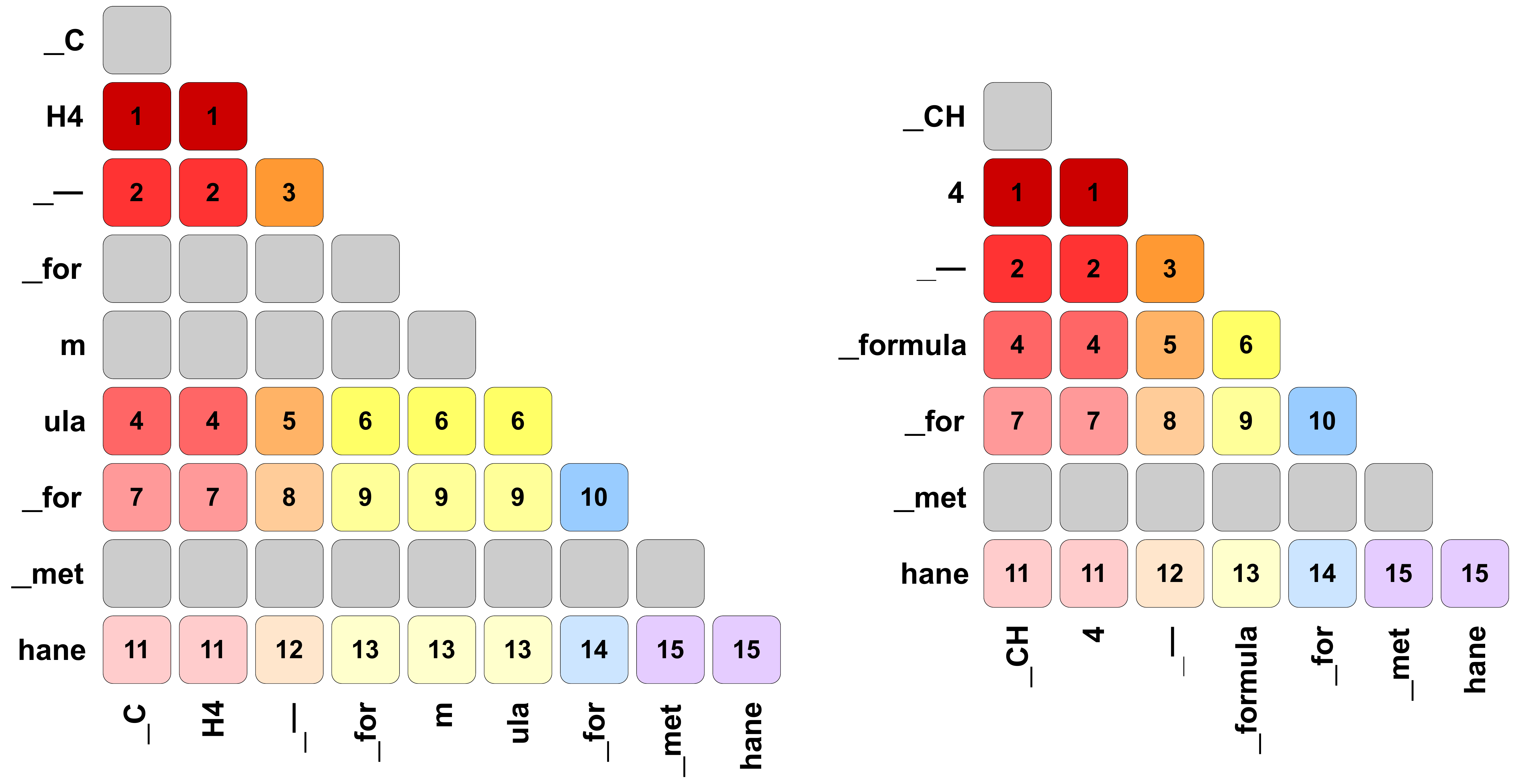}
    \caption{Sequence-level visualization of the AIM alignment between teacher and student models for the same text as Figure~\ref{fig:aim}. Each cell of the (causally masked, hence triangular) attention matrix is the weighted value state $\vv^*_{i,j}$ for query token $t_i$ (rows) and key/value token $t_j$ (columns); the left matrix uses the original tokenizer $T$ and the right matrix the new tokenizer $T^{\prime}$. Numbers and matching colors within a matrix identify tokens that are aggregated into the same segment-level state $\bm{\mathfrak{s}}_{i, k}$, and numbers and colors across the two matrices indicate corresponding pairs $\bm{\mathfrak{s}}_{\ell_T(i), j}$ and $\bm{\mathfrak{s}}^{\prime}_{\ell_{T^{\prime}}(i), j}$ that are matched by the loss $\mathcal{L}^*$.}
    \label{fig:tokenization_matrix}
\end{figure}

\clearpage

%%%%%%%%%%%%%%%%%%%%%%%%%%%%%%%%%%%%%%%%%%%%%%%%%%%%%%%%%%%%

% \newpage
% \input{checklist.tex}

\end{document}